\documentclass[10pt,twocolumn,letterpaper]{article}

\usepackage[table,xcdraw]{xcolor}

\usepackage{wacv}              

\usepackage[accsupp]{axessibility}
\usepackage{times}
\usepackage{epsfig}
\usepackage{graphicx}
\usepackage{amsmath}
\usepackage{amssymb}
\usepackage{booktabs}
\usepackage{multirow}
\usepackage{subcaption}
\usepackage{cuted}
\usepackage{appendix}
\usepackage{rotating}
\DeclareMathOperator*{\argmax}{arg\,max}

\newtheorem{definition}{Definition}
\makeatletter
\renewcommand\paragraph{\@startsection{paragraph}{4}{\z@}{1ex \@plus1ex \@minus.2ex}{-1em}{\normalfont\normalsize\bfseries}}
\makeatother

\usepackage[pagebackref,breaklinks,colorlinks]{hyperref}

\usepackage[capitalize]{cleveref}
\crefname{section}{Sec.}{Secs.}
\Crefname{section}{Section}{Sections}
\Crefname{table}{Table}{Tables}
\crefname{table}{Tab.}{Tabs.}
\usepackage{twemojis}

\def\wacvPaperID{1074} 
\def\confName{WACV}
\def\confYear{2024}

\begin{document}

\title{SOAP: Cross-sensor Domain Adaptation for 3D Object Detection Using Stationary Object Aggregation Pseudo-labelling}

\author{Chengjie Huang \qquad Vahdat Abdelzad \qquad Sean Sedwards \qquad Krzysztof Czarnecki \\
University of Waterloo\\
{\tt\small \{c.huang,vahdat.abdelzad,sean.sedwards, k2czarne\}@uwaterloo.ca}}

 \maketitle
\begin{abstract}
We consider the problem of cross-sensor domain adaptation in the context of LiDAR-based 3D object detection and propose Stationary Object Aggregation Pseudo-labelling (SOAP) to generate high quality pseudo-labels for stationary objects. In contrast to the current state-of-the-art in-domain practice of aggregating just a few input scans, SOAP aggregates entire sequences of point clouds at the input level to reduce the sensor domain gap. Then, by means of what we call \emph{quasi-stationary training} and \emph{spatial consistency post-processing}, the SOAP model generates accurate pseudo-labels for stationary objects, closing a minimum of 30.3\% domain gap compared to few-frame detectors. Our results also show that state-of-the-art domain adaptation approaches can achieve even greater performance in combination with SOAP, in both the unsupervised and semi-supervised settings.

\end{abstract}

\section{Introduction}\label{sec:introduction}

\begin{figure}[t]
    \centering
    \vspace{2em}
    \begin{subfigure}[t]{0.3\columnwidth}
        \includegraphics[width=\textwidth]{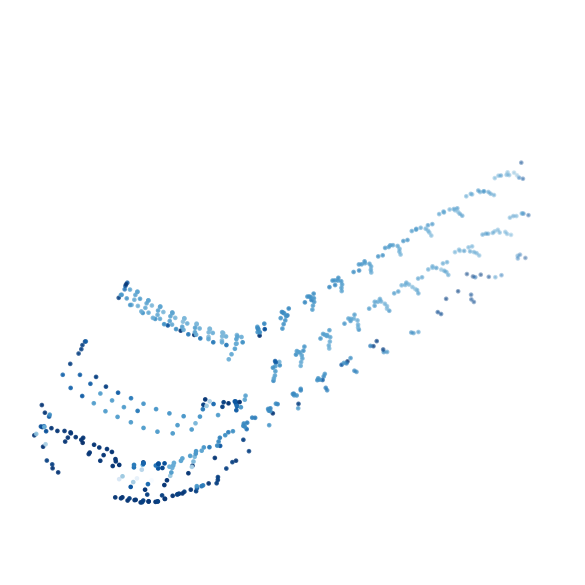}
        \caption{nuScenes sparse}
        \label{fig:nuScenes_sparse}
    \end{subfigure}
    \begin{subfigure}[t]{0.3\columnwidth}
        \includegraphics[width=\textwidth]{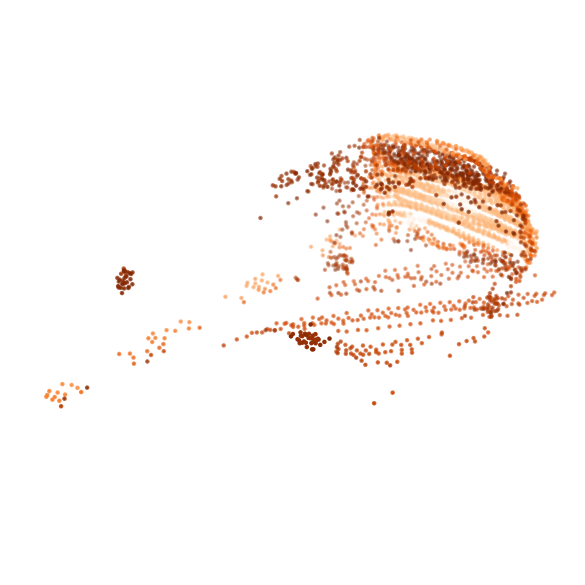}
        \caption{Waymo sparse}
        \label{fig:Waymo_sparse}
    \end{subfigure}
    \begin{subfigure}[t]{0.38\columnwidth}
        \includegraphics[width=\textwidth]{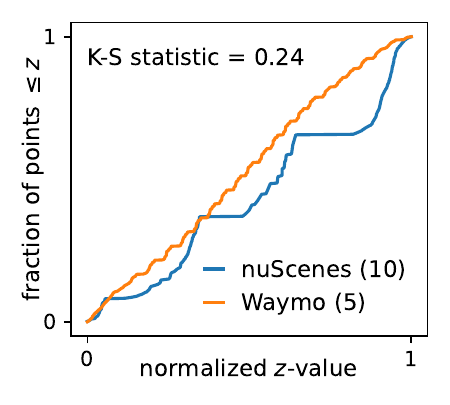}
        \caption{CDFs for sparse}
        \label{fig:CDF_sparse}
    \end{subfigure}
    \begin{subfigure}[t]{0.3\columnwidth}
        \includegraphics[width=\textwidth]{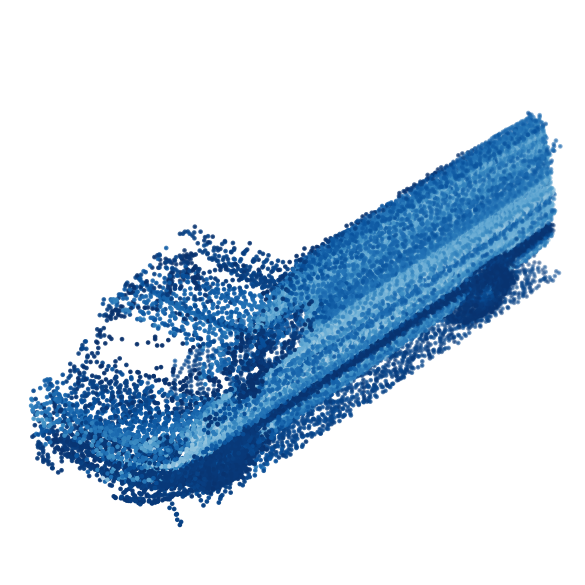}
        \caption{nuScenes dense}
        \label{fig:nuScenes_dense}
    \end{subfigure}
    \begin{subfigure}[t]{0.3\columnwidth}
        \includegraphics[width=\textwidth]{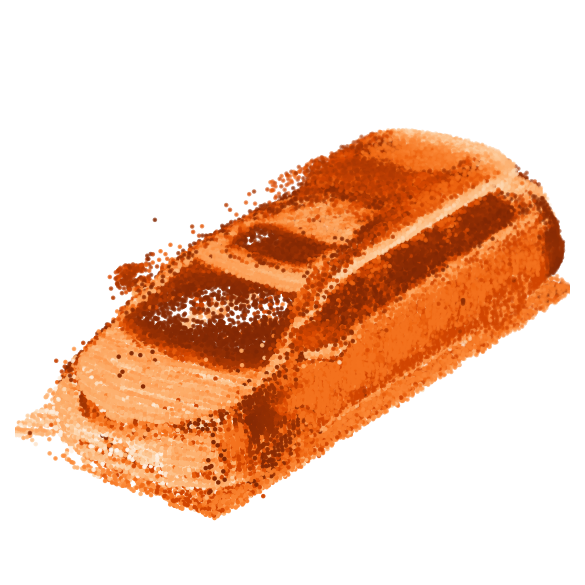}
        \caption{Waymo dense}
        \label{fig:Waymo_dense}
    \end{subfigure}
        \begin{subfigure}[t]{0.38\columnwidth}
        \includegraphics[width=\textwidth]{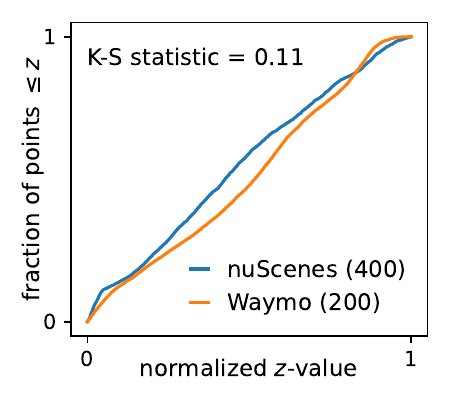}
        \caption{CDFs for dense}
        \label{fig:CDF_dense}
    \end{subfigure}
    \caption{Scan lines are evident in point clouds when only few input frames are used \subref{fig:nuScenes_sparse}\subref{fig:Waymo_sparse}, appearing as obvious modes in CDF plots that largely differ because of the modes \subref{fig:CDF_sparse}. Aggregating many more frames removes visible scan lines \subref{fig:nuScenes_dense}\subref{fig:Waymo_dense} and makes CDFs for similar objects in different datasets more alike \subref{fig:CDF_dense}}
    \label{fig:cross-sensor}
\end{figure}

LiDAR sensors are commonly used in autonomous driving and other safety-critical robotic applications to provide accurate 3D localization of objects. State-of-the-art (SOTA) LiDAR-based object detectors currently use deep neural networks trained via supervised learning, requiring a large amount of realistic data labelled by human annotators. Annotation is expensive, motivating the re-use of existing labelled datasets, but these typically have a limited \emph{domain}, e.g., a specific sensor configuration and relatively few geographic locations and weather conditions. It is well known that detectors trained with a dataset from one domain experience a significant degradation of performance when fed data from a different domain.

In this work, we tackle the cross-sensor domain adaptation problem that arises whenever a detector is required to interpret data from a sensor different to that on which it was trained. Specifically, given a detector trained on labelled point clouds collected using one sensor (the \emph{source domain}), we aim to improve the detector's performance on point clouds collected using a different sensor (the \emph{target domain}), either with no labels (unsupervised) or with only a small number of labels from the target domain (semi-supervised).
This situation arises commonly when a LiDAR sensor is updated or a fleet of autonomous vehicles uses multiple sensors.

Cross-sensor domain adaptation can present a formidable challenge because similar objects scanned by different LiDAR sensors may have very different scan patterns, even after the widely-adopted few-frame input aggregation. \Cref{fig:nuScenes_sparse,fig:Waymo_sparse} show point clouds of vehicles at similar distances from their respective sensors, created by aggregating 10 frames from nuScenes~\cite{caesar2020nuscenes} and 5 frames from Waymo~\cite{sun2020waymo} datasets, respectively. There are evident scan lines and obvious visual dissimilarities between the point clouds. There are also significant differences in the corresponding cumulative density function (CDF) plots of the $z$-component of point positions (\cref{fig:CDF_sparse}).

Current SOTA 3D object detectors, while achieving impressive in-domain performance, still have a considerable performance gap when they are applied to cross-sensor point clouds, whether using single- or few-frame input. This is demonstrated in~\Cref{tab:few-frame-domain-gap}, where we compare the performance of a VoxelNeXt~\cite{chen2023voxelnext} model trained and evaluated on the nuScenes~\cite{caesar2020nuscenes} and Waymo~\cite{sun2020waymo} datasets.

We attribute this drop in performance in large part to the different scan patterns mentioned above. This view is supported by a recent study based on simulation~\cite{fang2023lidarcs} that suggests the difference in scan patterns alone can have a substantial impact on the performance of existing object detectors.
At the same time, we observe that aggregating many more frames tends to reduce the scan patterns. This is illustrated in~\cref{fig:nuScenes_dense,fig:Waymo_dense}, which show dense point clouds created by aggregating 400 nuScenes and 200 Waymo frames, respectively. There are no visible scan lines and the multi-modality evident with few frames has disappeared from the CDF plots in~\cref{fig:CDF_dense}.

SOTA methods in cross-sensor domain adaptation often employ pseudo-labelling, where a model trained on labelled data is used to generate labels for unlabelled data. Various approaches have been proposed to improve pseudo-label quality and regularize training with pseudo-labels~\cite{yang2021st3d,yang2021st3d++,Wang2022SSDA3DSD}, but these approaches do not appear to explicitly address the important difference in scan-patterns between different domains.

Given all of the above, we propose \emph{Stationary Object Aggregation Pseudo-labelling} (SOAP) to improve cross-sensor pseudo-label accuracy by exploiting scene-level full-sequence aggregation of input point clouds to close the domain gap caused by scan-patterns.

SOAP uses sequential point clouds produced under realistic driving conditions by existing LiDAR sensors that are widely used in autonomous driving systems. It enhances existing pre-trained detectors, improving their stationary object performance while retaining dynamic object pseudo-labels. SOAP pseudo-labels can be used for updating detectors or bootstrapping annotations.

SOAP is motivated by the facts that (i) aggregation improves the representation of sparsely-scanned objects~\cite{wang2022sparse2dense}, (ii) sensor-specific scan patterns are reduced by full-sequence aggregation (\cref{fig:cross-sensor}), and (iii) stationary objects respond well to full-sequence aggregation and are a statistically important component of object detection: at least two thirds of cars are stationary at some point in sequences in major realistic driving datasets~\cite{caesar2020nuscenes,sun2020waymo,kesten2019lyft,wilson2021argoverse}.

Extensive experiments using nuScenes and Waymo datasets show SOAP pseudo-labels can close a minimum of 30.3\% overall domain gap compared to few-frame detectors without access to any target domain labels. SOAP also complements other SOTA domain adaptation methods, including ST3D~\cite{yang2021st3d} and SSDA3D~\cite{Wang2022SSDA3DSD}, improving their already impressive results in both unsupervised and semi-supervised settings. In nuScenes $\rightarrow$ Waymo setting using CenterPoint~\cite{yin2021centerpoint}, for example, using SOAP closes 42.6\% domain gap compared to the 9.5\% closed by ST3D. With only 1\% target domain labels, SOAP closes 86.8\% domain gap compared to the 81.4\% closed by SSDA3D.

\begin{table}[t]
\centering
\footnotesize
\setlength{\tabcolsep}{0.4em}
\renewcommand{\arraystretch}{0.9}
\begin{tabular}{cccc}
\toprule
 & \textbf{\# frames} & \textbf{AP} & $\Delta$ \\ \midrule
\multirow{2}{*}{nuScenes $\rightarrow$ Waymo} & 1 & 7.0 & \textbf{{\scriptsize --70.5}}   \\
 & 5 & 20.4 & \textbf{{\scriptsize --57.1}}   \\ \midrule
Waymo in-domain & 5 & 77.5 &  \\ \midrule\midrule
\multirow{2}{*}{Waymo $\rightarrow$ nuScenes} & 1 &  42.7 & \textbf{{\scriptsize --40.8}} \\
 & 10 & 49.0 & \textbf{{\scriptsize --34.5}}  \\ \midrule
nuScenes in-domain & 10 & 83.5 & \\ \bottomrule
\end{tabular}
\vspace{-0.5em}
\caption{Degradation ($\Delta$) w.r.t. in-domain performance of SOTA detector VoxelNeXt~\cite{chen2023voxelnext} in Waymo $\leftrightarrow$ nuScenes cross-sensor setting. Increasing the number of aggregated frames improves over single-frame input, but there is still a substantial performance gap.}
\label{tab:few-frame-domain-gap}
\end{table}

Our main contributions are as follows:
\begin{itemize}
    \item We propose SOAP to effectively utilize full-sequence scene-level aggregation and exploit the properties of the pseudo-labels.
    \item We demonstrate that full-sequence scene-level aggregation, though not optimal for in-domain settings, can be used to improve cross-sensor performance.
    \item We conduct extensive experiments to demonstrate SOAP's high quality pseudo-labels and synergy with SOTA domain adaptation methods.
\end{itemize}


\section{Related Work} \label{sec:related-work}
\begin{figure*}
    \centering
    \includegraphics[width=0.85\linewidth]{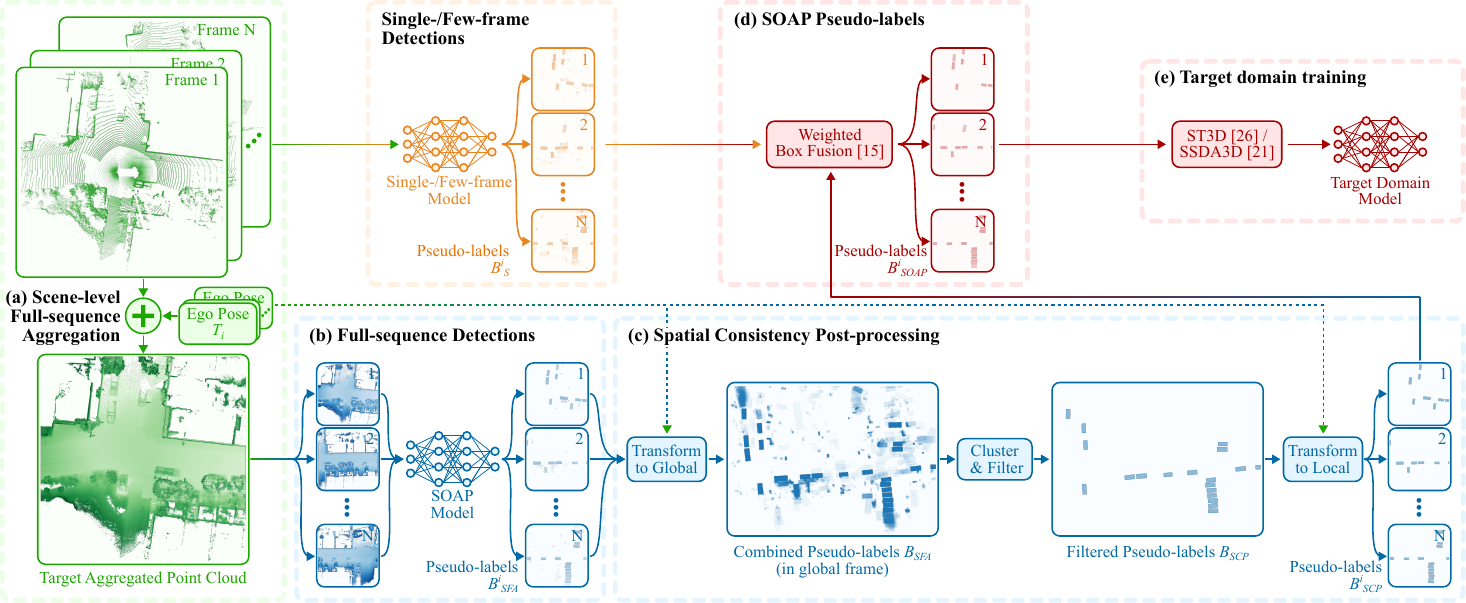}
    \caption{Overview of Stationary Object Aggregation Pseudo-labelling (SOAP) pipeline. (a) We first perform Scene-level Full-sequence Aggregation (SFA) using pose transforms. (b) We propose Quasi-Stationary Training (QST) to train a SOAP model to detect stationary objects. (c) The predictions are refined via Spatial Consistency Post-processing (SCP). (d) The predictions from a pre-trained single-/few-frame detector and the SOAP model are combined using Weighted Box Fusion (WBF)~\cite{solovyev2021wbf}. (e) The final SOAP pseudo-labels can be used in combination with SOTA methods to fine-tune a target domain detector.}
    \label{fig:soap}
\end{figure*}

\paragraph{Domain adaptation for 3D object detection:}
Many methods have been proposed to address the domain adaptation problem for 3D object detection. One line of work involves improving model robustness via regularization~\cite{wang2022sparse2dense,xu2021spg}. Other works attempt to close the domain gap via domain mapping~\cite{barrera2021bevcyclegan,corral2021lcp} or input~\cite{wang2020germany,tsai2022see,tsai2022seevcn}, feature~\cite{yihan20213dcoco,zhang2021srdan,luo2021mlcnet}, and output~\cite{ding2022jst} alignment. To take advantage of available target domain data, SOTA methods often use pseudo-labelling. Pseudo-labels can be improved via tracking-based refinement~\cite{you2022dreaming,fruhwirth2021fast3d} or iterative self-training~\cite{yang2021st3d,yang2021st3d++}. When a small amount of target labels are available, CutMix and MixUp have also been shown to be effective techniques to incorporate labelled target data~\cite{Wang2022SSDA3DSD}. As we will show, SOAP is parallel to and can complement existing work in this area.

\paragraph{Offline pseudo-labelling:}
Previous studies have shown that increasing the number of frames aggregated at scene-level leads to diminishing returns~\cite{caesar2020nuscenes} or even performance degradation~\cite{chen2022mppnet}, especially for dynamic objects~\cite{yang20213dman}. As a result, SOTA offline pseudo-labelling methods use a single- or few-frame detector to generate initial predictions, followed by offline tracking and a second stage refinement that utilizes full-sequence point clouds aggregated at object- or track-level~\cite{yang2021auto4d,qi2021offboard,fan2023ctrl,ma2023detzero}. Offline pseudo-labelling has achieved impressive in-domain results, even surpassing human performance, but they have not yet been explored in cross-sensor setting. SOAP, on the other hand, takes a completely different view from the aforementioned works and directly uses scene-level aggregated point clouds as input to provide better pseudo-labels than single- or few-frame detectors for cross-sensor domain adaptation setting.


\section{Our approach: SOAP} \label{sec:soap}
In this section, we describe the details of \emph{Stationary Object Aggregation Pseudo-labelling} (SOAP). The main components of SOAP are: (i) \emph{Scene-level Full-sequence Aggregation} (SFA), which produces aggregated point clouds from the entire input sequence; (ii) \emph{Quasi-Stationary Training} (QST) that is used to train a pseudo-labelling model to detect stationary objects; and (iii) \emph{Spatial Consistency Post-processing} (SCP) that enhances pseudo-labels by exploiting the stationarity of the predictions. SOAP is used in combination with a pre-trained model to generate high quality pseudo-labels for stationary objects while retaining dynamic object pseudo-labels. An overview of the SOAP pipeline is shown in~\cref{fig:soap}.

\subsection{Scene-level Full-sequence Aggregation}  \label{sec:sa}

\emph{Scene-level full-sequence aggregation} (SFA) involves projecting a sequence of point clouds to a global coordinate system, where the point clouds are concatenated into a single dense point cloud, as illustrated in~\cref{fig:soap}a.
Formally, given a sequence of point clouds $P_1, P_2,\dots, P_N$, where $P_i = \{p_i^1,p_i^2,\dots,p_i^{M_i}\}\subset\mathbb{R}^3$, and corresponding sequence of $\mathbb{SE}(3)$ pose transformations $T_1,T_2,\dots,T_N$, which transform the point clouds from the local LiDAR or vehicle coordinate system to a common global coordinate system, the point cloud aggregation process in the global coordinate system is defined by
\begin{equation}
    P^* = \bigcup_{i=1,2,\dots,N} \left\{ T_ip_i^j \right\}_{j=1,2,\dots,M_i}.
\end{equation}
During training, for a given frame $i$, the aggregated point cloud $P^*$ is transformed back to the local coordinate system using the inverse pose transform $T_i^{-1}$.

Scene-level aggregation of a few frames in a short time window has been shown to be effective at producing denser input point clouds and consequently better detection performance~\cite{caesar2020nuscenes}. On the other hand, due to the diminishing returns~\cite{caesar2020nuscenes} and performance degradation~\cite{yang20213dman,chen2022mppnet} observed for scene-level aggregation with large temporal windows, full-sequence aggregation has only been attempted at object- and track-level~\cite{yang2021auto4d,qi2021offboard,fan2023ctrl,ma2023detzero}.

Compared to few-frame aggregation, SFA increases point density and provides richer geometric information for stationary objects. In addition, unlike object- and track-level aggregation used by prior work, SFA does not rely on object annotations or initial predictions produced by single- or few-frame detectors, which we show are unreliable in the cross-sensor setting~\Cref{tab:few-frame-domain-gap}. SFA is therefore very suitable for its proposed application, since $P^*$ can be obtained for both the source and (unlabelled) target domains.

\begin{figure}
    \centering
    \includegraphics{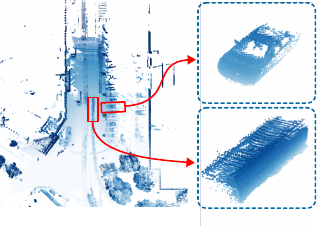}
    \vspace{-1em}
    \caption{Example of a point cloud generated by SFA. Dynamic objects are distorted while stationary objects are densified.}
    \label{fig:fsa}
\end{figure}

SFA tends to distort dynamic object point clouds, due to the motion of the objects not being corrected during aggregation. This is depicted in \cref{fig:fsa}. By contrast, stationary objects are densified with a more complete and accurate geometry compared to single- or few-frame point clouds. As noted above and in~\cref{fig:cross-sensor}, the aggregation process also weakens the LiDAR-specific scan patterns.
\subsection{Quasi-Stationary Training} \label{sec:qst}

Although annotations are available for the source domain, training a model to detect stationary objects from aggregated point clouds is not straightforward. A naive approach is by filtering the ground truth annotations during training based on speed estimates or the overall displacement of the object.
However, since the speed of an object can change over time, especially for sequences that span a large time window, the object can be near stationary during part of the sequence and moving in other parts. If the majority of the observed points come from the part of the sequence where the object is near stationary, the object's aggregated point cloud will have little distortion, as if the object were stationary for the entirety of the sequence.

\begin{figure}
    \centering
    \begin{subfigure}[b]{0.38\columnwidth}
        \includegraphics[width=\textwidth]{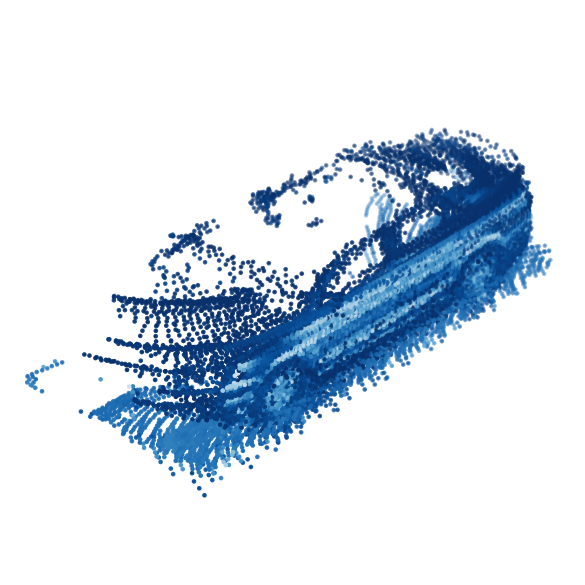}
        \caption{Aggregated point cloud}
    \end{subfigure}
        \begin{subfigure}[b]{0.38\columnwidth}
        \includegraphics[width=\textwidth]{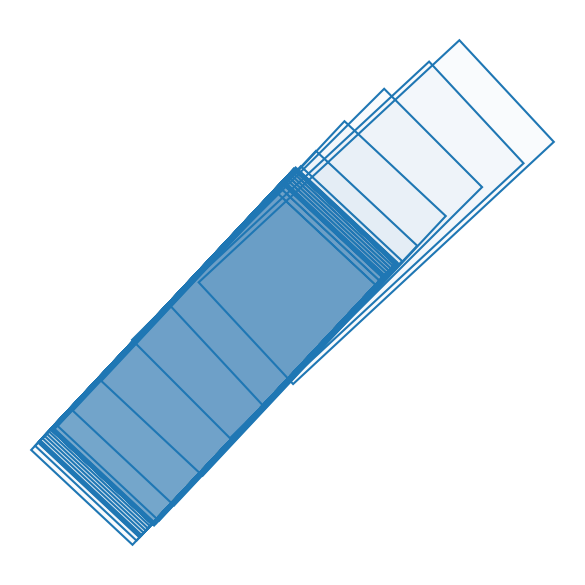}
        \caption{Object trajectory (BEV)}
    \end{subfigure}
    \vspace{-0.5em}
    \caption{Example of a quasi-stationary object. This object reached a maximum speed of 1.4\,m/s with a total displacement of 3.9\,m, and thus would be eliminated by naive filtering.}
    \label{fig:qs-object}
\end{figure}

We refer to such objects as \emph{quasi-stationary} objects.
\Cref{fig:qs-object} depicts an example of such an object, which would be excluded based on a naive speed or displacement criterion. Doing so would result in undistorted objects remaining in the aggregated point clouds, but without a positive label, causing confusion and reducing model performance.

To avoid excluding these quasi-stationary objects, we formally define the notion of quasi-stationarity using a \emph{quasi-stationary score} (QSS) that takes into account both the movement of the objects and how much each observation contributes to the final aggregated point clouds.

\begin{definition}[QSS]
Let $\{b_1,b_2,\dots,b_N\}$ be a set of $N$ bounding boxes of an object annotated in a sequence within a common coordinate system and $\mathrm{C}(b_i)$ be the number of points observed in the bounding box $b_i$. For a given bounding box observation $b_i$, $\mathrm{QSS}$ is defined as the average IoU between $b_i$ and other bounding boxes $b_j$, weighted by the fractions of points contributed by $b_j$:
\begin{equation}
    \mathrm{QSS}(b_i) = \sum_{j=1}^N \frac{\mathrm{C}(b_j)}{\sum_{k=1}^N \mathrm{C}(b_k)} \mathrm{IoU} (b_i, b_j)
\end{equation}
\end{definition}

Intuitively, the $\mathrm{QSS}$ can be interpreted as how likely it is that the point cloud for a given object is undistorted at the location of $b_i$. For example, if another observation $b_j$ has little overlap with $b_i$ (indicating object movement) but contains only a few points, then the final aggregated point cloud is not likely to be distorted by $b_j$. Alternatively, if $b_j$ has a large overlap with $b_i$ and also contains a large fraction of points, then the object is likely to undistorted at location $b_i$.

Finally, the most likely location $b^*$ of the object in the aggregated point clouds and the degree $s^*$ of the point cloud being free from distortion can be estimated as follows:
\begin{align}
    b^* &= {\argmax}_{i}\; \mathrm{QSS}(b_i)\\
    s^* &= {\max}_{i} \; \mathrm{QSS}(b_i)
\end{align}
We refer to $b^*$ as the quasi-stationary bounding box and $s^*$ as the corresponding QSS.
Objects with a large QSS $s^*>\epsilon$ for some $\epsilon$ can be considered quasi-stationary.
For instance, the object in \Cref{fig:qs-object} has QSS $s^*=0.91$.

\paragraph{Out-of-sight quasi-stationary objects:}
In the labelled source domain dataset, we notice objects not visible from the current frame are sometimes not labelled. Since SFA utilizes the entire sequence to create a dense representation of the scene, out-of-sight quasi-stationary objects will also be densified in the corresponding aggregated point cloud. This creates inconsistent labels where in a given frame, some dense objects have annotations while others do not. To address this problem, we construct spatially consistent training labels by projecting the quasi-stationary bounding boxes $b^*$ to \emph{all} frames in the sequence, even if the object is not observed in some frames.

\subsection{Spatial Consistency Post-processing} \label{sec:scp}

Since the SOAP model is tasked to detect only quasi-stationary objects from aggregated point clouds, a detected object should have consistent predictions in the global coordinate system across multiple frames. To utilize this stationarity property of the pseudo-labels, we propose \emph{Spatial Consistency Post-processing} (SCP) to eliminate false positive predictions and recover false negative objects, improving pseudo-label quality.

As illustrated in~\cref{fig:soap}c, SCP is performed by obtaining per-frame predicted bounding boxes (denoted by $B^i_\mathit{SFA}$ for frame $i$) using the SOAP model and then gathering all predictions in the global coordinate system. Gathering all predictions requires transforming per-frame predictions (bounding boxes) using the corresponding ego pose transformation $T_i$:
\begin{equation}
    B_\mathit{SFA} = \bigcup T_i B_\mathit{SFA}^i
\end{equation}

The bounding boxes $B_\mathit{SFA}$ are clustered based on an IoU threshold $\mu$. To ensure the pseudo-labels are consistent across multiple frames, we eliminate the clusters with a number of detections fewer than threshold $\eta$ that depends on the frame rate of the dataset. 

For the remaining clusters, the boxes in each cluster are combined into a single bounding box per cluster, similar to Weighted Boxes Fusion (WBF)~\cite{solovyev2021wbf}. Specifically, we use the heading $\theta$ of the most confident prediction in each cluster and average other attributes, including position ($x,y,z$), size $(w,l,h)$ and velocity $(v_x, v_y)$, weighted by the confidence of each box. Finally, non-maximum suppression is applied to remove any overlapping predictions in the global coordinate system. The final bounding boxes in the global coordinate system are denoted by $B_\mathit{SCP}$.

To obtain the pseudo-labels $B_\mathit{SCP}^i$ for each frame $i$ in the sequence, we transform $B_\mathit{SCP}$ back to each frame's local coordinate system using the inverse pose transformation $T_i^{-1}$, defined as follows:
\begin{equation}
    B_\mathit{SCP}^i = T_i^{-1} B_\mathit{SCP}
\end{equation}

Since an object may be occluded in a frame and have very few or no points in the sparse point cloud, we remove any bounding boxes that contain no points, so that the pseudo-labels are reasonable with respect to the frame's sparse point cloud.

\subsection{SOAP pseudo-labels}
In order to recover a complete set of pseudo-labels for both stationary and dynamic objects, SOAP utilizes the predictions from a pre-trained single- or few-frame detector capable of detecting dynamic objects. The SOAP and pre-trained models are calibrated separately with Beta Calibration~\cite{kull2017beta} using source domain data.
Let $B_S^i$ be the bounding boxes predicted by the pre-trained detector for frame $i$, then the SOAP pseudo-labels, denoted by $B_\mathit{SOAP}^i$, are obtained by combining $B_S^i$ and $B_\mathit{SCP}^i$ using WBF~\cite{solovyev2021wbf}.

Our results show that this simple approach can improve existing sparse pseudo-labels by a large margin. We leave it as future work to study more optimal ways of combining pseudo-labels or obtaining dynamic pseudo-labels directly from aggregated point clouds.

\section{Experiments} \label{sec:experiments}
\begin{table*}[ht]
\centering
\footnotesize
\setlength{\tabcolsep}{0.4em}
\resizebox{0.85\textwidth}{!}{%
\begin{tabular}{cccllllllllllll}
\toprule
\multirow{2}{*}{\textbf{Architecture}} & \multirow{2}{*}{\textbf{Method}} & \multirow{2}{*}{\textbf{\begin{tabular}[c]{c}Training\\ Data\end{tabular}}} & \multicolumn{4}{c}{\textbf{Overall}} & \multicolumn{4}{c}{\textbf{0--30\,m}} & \multicolumn{4}{c}{\textbf{30--50\,m}} \\
 &  &  & \multicolumn{2}{l}{\textbf{Level 1}} & \multicolumn{2}{l}{\textbf{Level 2}} & \multicolumn{2}{l}{\textbf{Level 1}} & \multicolumn{2}{l}{\textbf{Level 2}} & \multicolumn{2}{l}{\textbf{Level 1}} & \multicolumn{2}{l}{\textbf{Level 2}} \\ \midrule
\multirow{5}{*}{CenterPoint~\cite{yin2021centerpoint}} & Direct & \multirow{2}{*}{$\{\mathcal{S}\}$} & 23.5 &  & 20.2 &  & 49.3 &  & 48.3 &  & 12.0 &  & 10.5 &  \\
 & SOAP (ours) &  & \textbf{50.9} & \textbf{{\scriptsize +51.0\%}} & \textbf{45.4} & \textbf{{\scriptsize +51.2\%}} & \textbf{69.4} & \textbf{{\scriptsize +47.0\%}} & \textbf{68.4} & \textbf{{\scriptsize +47.1\%}} & \textbf{47.5} & \textbf{{\scriptsize +55.5\%}} & \textbf{43.6} & \textbf{{\scriptsize +55.5\%}} \\ \cmidrule(l){2-15} 
 & ST3D~\cite{yang2021st3d} & \multirow{2}{*}{$\{\mathcal{S}, \mathcal{T}_P\}$} & 28.6 & {\scriptsize +9.5\%} & 24.6 & {\scriptsize +8.9\%} & 56.5 & {\scriptsize +16.8\%} & 55.3 & {\scriptsize +16.4\%} & 18.7 & {\scriptsize +10.5\%} & 16.5 & {\scriptsize +10.1\%} \\
 & ST3D + SOAP (ours) &  & \textbf{46.4} & \textbf{{\scriptsize +42.6\%}} & \textbf{40.6} & \textbf{{\scriptsize +41.5\%}} & \textbf{68.6} & \textbf{{\scriptsize +45.1\%}} & \textbf{67.5} & \textbf{{\scriptsize +45.0\%}} & \textbf{41.8} & \textbf{{\scriptsize +46.6\%}} & \textbf{37.6} & \textbf{{\scriptsize +45.5\%}} \\ \cmidrule(l){2-15} 
 & Oracle & $\{\mathcal{T}\}$ & 77.2 &  & 69.4 &  & 92.1 &  & 91.0 &  & 76.0 &  & 70.1 &  \\ \midrule\midrule
\multirow{5}{*}{VoxelNeXt~\cite{chen2023voxelnext}} & Direct & \multirow{2}{*}{$\{\mathcal{S}\}$} & 20.4 &  & 17.5 &  & 44.2 &  & 43.3 &  & 9.5 &  & 8.4 &  \\
 & SOAP (ours) &  & \textbf{50.9} & \textbf{{\scriptsize +53.4\%}} & \textbf{45.6} & \textbf{{\scriptsize +53.8\%}} & \textbf{67.5} & \textbf{{\scriptsize +48.5\%}} & \textbf{66.4} & \textbf{{\scriptsize +48.3\%}} & \textbf{48.6} & \textbf{{\scriptsize +58.5\%}} & \textbf{44.7} & \textbf{{\scriptsize +58.6\%}} \\ \cmidrule(l){2-15} 
 & ST3D~\cite{yang2021st3d} & \multirow{2}{*}{$\{\mathcal{S}, \mathcal{T}_P\}$} & 35.0 & {\scriptsize +25.6\%} & 30.2 & {\scriptsize +25.3\%} & 60.7 & {\scriptsize +34.4\%} & 59.5 & {\scriptsize +33.9\%} & 27.9 & {\scriptsize +27.5\%} & 24.8 & {\scriptsize +26.5\%} \\
 & ST3D + SOAP (ours) &  & \textbf{45.6} & \textbf{{\scriptsize +44.1\%}} & \textbf{40.0} & \textbf{{\scriptsize +43.1\%}} & \textbf{65.9} & \textbf{{\scriptsize +45.2\%}} & \textbf{64.8} & \textbf{{\scriptsize +45.0\%}} & \textbf{41.8} & \textbf{{\scriptsize +48.4\%}} & \textbf{37.7} & \textbf{{\scriptsize +47.3\%}} \\ \cmidrule(l){2-15} 
 & Oracle & $\{\mathcal{T}\}$ & 77.5 &  & 69.7 &  & 92.2 &  & 91.1 &  & 76.3 &  & 70.3 &  \\ \toprule
\multicolumn{15}{c}{$\mathcal{S}$: labelled source domain; $\mathcal{T}$: labelled target domain; $\mathcal{T}_P$: pseudo-labelled target domain}
\end{tabular}
}
\vspace{-0.5em}
\caption{Unsupervised domain adaptation results for nuScenes $\rightarrow$ Waymo, where Waymo dataset is unlabelled. The percentages represent the amount of the Direct--Oracle domain gap closed.}
\label{tab:st3d-nus-waymo}
\end{table*}
\begin{table*}[ht]
\centering
\footnotesize
\setlength{\tabcolsep}{0.4em}
\resizebox{0.85\textwidth}{!}{%
\begin{tabular}{cccllllllllllll}
\toprule
\multirow{2}{*}{\textbf{Architecture}} & \multirow{2}{*}{\textbf{Method}} & \multirow{2}{*}{\textbf{\begin{tabular}[c]{c}Training\\ Data\end{tabular}}} & \multicolumn{4}{c}{\textbf{Overall}} & \multicolumn{4}{c}{\textbf{0--30\,m}} & \multicolumn{4}{c}{\textbf{30--50\,m}} \\
 &  &  & \multicolumn{2}{l}{\textbf{mAP}} & \multicolumn{2}{l}{\textbf{NDS}} & \multicolumn{2}{l}{\textbf{mAP}} & \multicolumn{2}{l}{\textbf{NDS}} & \multicolumn{2}{l}{\textbf{mAP}} & \multicolumn{2}{l}{\textbf{NDS}} \\ \midrule
\multirow{5}{*}{CenterPoint~\cite{yin2021centerpoint}} & Direct & \multirow{2}{*}{$\{\mathcal{S}\}$} & 51.7 &  & 69.6 &  & 67.6 &  & 78.9 &  & 27.6 &  & 49.7 &  \\
 & SOAP (ours) &  & \textbf{61.4} & \textbf{{\scriptsize +30.3\%}} & \textbf{76.9} & \textbf{{\scriptsize +39.2\%}} & \textbf{73.1} & \textbf{{\scriptsize +21.7\%}} & \textbf{83.9} & \textbf{{\scriptsize +33.3\%}} & \textbf{41.9} & \textbf{{\scriptsize +40.1\%}} & \textbf{64.4} & \textbf{{\scriptsize +60.2\%}} \\ \cmidrule(l){2-15} 
 & ST3D~\cite{yang2021st3d} & \multirow{2}{*}{$\{\mathcal{S}, \mathcal{T}_P\}$} & 59.3 & {\scriptsize +23.8\%} & 72.9 & {\scriptsize +17.7\%} & \textbf{74.2} & \textbf{{\scriptsize +26.1\%}} & 82.4 & {\scriptsize +23.3\%} & 34.1 & {\scriptsize +18.2\%} & 52.7 & {\scriptsize +12.3\%} \\
 & ST3D + SOAP (ours) &  & \textbf{61.5} & \textbf{{\scriptsize +30.6\%}} & \textbf{75.4} & \textbf{{\scriptsize +31.2\%}} & 73.9 & {\scriptsize +24.9\%} & \textbf{83.0} & \textbf{{\scriptsize +27.3\%}} & \textbf{42.5} & \textbf{{\scriptsize +41.7\%}} & \textbf{60.4} & \textbf{{\scriptsize +43.9\%}} \\ \cmidrule(l){2-15} 
 & Oracle & $\{\mathcal{T}\}$ & 83.7 &  & 88.2 &  & 92.9 &  & 93.9 &  & 63.3 &  & 74.1 &  \\ \midrule\midrule
\multirow{5}{*}{VoxelNeXt~\cite{chen2023voxelnext}} & Direct & \multirow{2}{*}{$\{\mathcal{S}\}$} & 49.0 &  & 68.3 &  & 62.5 &  & 76.5 &  & 28.8 &  & 50.6 &  \\
 & SOAP (ours) &  & \textbf{61.5} & \textbf{{\scriptsize +36.2\%}} & \textbf{77.0} & \textbf{{\scriptsize +44.2\%}} & \textbf{72.5} & \textbf{{\scriptsize +32.5\%}} & \textbf{83.5} & \textbf{{\scriptsize +40.5\%}} & \textbf{43.9} & \textbf{{\scriptsize +45.5\%}} & \textbf{65.3} & \textbf{{\scriptsize +65.3\%}} \\ \cmidrule(l){2-15} 
 & ST3D~\cite{yang2021st3d} & \multirow{2}{*}{$\{\mathcal{S}, \mathcal{T}_P\}$} & 54.2 & {\scriptsize +15.1\%} & 70.8 & {\scriptsize +12.7\%} & \textbf{64.6} & \textbf{{\scriptsize +6.8\%}} & 77.8 & {\scriptsize +7.5\%} & 38.0 & {\scriptsize +27.7\%} & 55.3 & {\scriptsize +20.9\%} \\
 & ST3D + SOAP (ours) &  & \textbf{56.0} & \textbf{{\scriptsize +20.3\%}} & \textbf{72.7} & \textbf{{\scriptsize +22.3\%}} & \textbf{64.6} & \textbf{{\scriptsize +6.8\%}} & \textbf{78.3} & \textbf{{\scriptsize +10.4\%}} & \textbf{43.7} & \textbf{{\scriptsize +44.9\%}} & \textbf{61.2} & \textbf{{\scriptsize +47.1\%}} \\ \cmidrule(l){2-15} 
 & Oracle & $\{\mathcal{T}\}$ & 83.5 &  & 88.0 &  & 93.3 &  & 93.8 &  & 62.0 &  & 73.1 &  \\ \toprule
\multicolumn{15}{c}{$\mathcal{S}$: labelled source domain; $\mathcal{T}$: labelled target domain; $\mathcal{T}_P$: pseudo-labelled target domain}
\end{tabular}
}
\vspace{-0.5em}
\caption{Unsupervised domain adaptation results for Waymo $\rightarrow$ nuScenes, where nuScenes dataset is unlabelled. The percentages represent the amount of the Direct--Oracle domain gap closed.}
\label{tab:st3d-waymo-nus}
\end{table*}

SOAP is evaluated in both unsupervised and semi-supervised domain adaptation settings. This section presents the experimental setup, domain adaptation results, and ablation study.

\subsection{Datasets} \label{sec:datasets}
We evaluate SOAP using two large-scale autonomous driving datasets for 3D object detection: nuScenes~\cite{caesar2020nuscenes} and Waymo~\cite{sun2020waymo}.
In what follows, we use the syntax \emph{source domain dataset} $\rightarrow$ \emph{target domain dataset} to denote the training and testing setting, respectively. 

The NuScenes dataset~\cite{caesar2020nuscenes} contains 1,000 sequences of 20 seconds each, collected in Boston and Singapore. The vehicle is equipped with a single Velodyne HDL-32E 32-beam top-mounted rotating LiDAR operating at 20\,Hz, yielding ${\approx}400$ point cloud scans per sequence, from which 40 keyframes are selected uniformly and annotated. Unless specified otherwise, all models trained on the nuScenes dataset use 10 sweeps (--0.5\,s) as input.

The Waymo dataset~\cite{sun2020waymo} contains 1,150 sequences of 20 seconds each, collected in San Francisco, Mountain View, and Phoenix. The Waymo dataset uses a 5-sensor setup with a single proprietary 64-beam top-mounted rotating LiDAR operating at 10\,Hz, and four side-mounted close-range LiDARs. All point clouds (${\approx}200$) in each sequence are annotated with bounding boxes. Due to much higher annotation frequency compared to nuScenes, we use 20\% uniformly sampled frames for training. Consistent with the nuScenes models, all models trained on the Waymo dataset use 5 sweeps (--0.5\,s) as input.

In addition to the difference in the point cloud and annotation frequency, the nuScenes dataset annotates 23 classes with 8 attributes, 10 of which are used in the object detection task, whereas Waymo contains annotations for only Vehicle, Cyclist, and Pedestrian. Following previous work~\cite{wang2020germany,yang2021st3d,tsai2022see,tsai2022seevcn,you2022dreaming,yang2021st3d++,luo2021mlcnet,yihan20213dcoco}, we select the common vehicle/car class for training and evaluation for all experiments.

\subsection{Evaluation}
\begin{table*}[ht]
\centering
\footnotesize
\setlength{\tabcolsep}{0.4em}
\resizebox{0.8\textwidth}{!}{%
\begin{tabular}{ccllllllllllll}
\toprule
\multirow{2}{*}{\textbf{Method}} & \multirow{2}{*}{\textbf{\begin{tabular}[c]{c}Training\\ Data\end{tabular}}} & \multicolumn{4}{c}{\textbf{Overall}} & \multicolumn{4}{c}{\textbf{0--30\,m}} & \multicolumn{4}{c}{\textbf{30--50\,m}} \\
 &  & \multicolumn{2}{l}{\textbf{Level 1}} & \multicolumn{2}{l}{\textbf{Level 2}} & \multicolumn{2}{l}{\textbf{Level 1}} & \multicolumn{2}{l}{\textbf{Level 2}} & \multicolumn{2}{l}{\textbf{Level 1}} & \multicolumn{2}{l}{\textbf{Level 2}} \\ \midrule
Direct & $\{\mathcal{S}\}$ & 23.5 &  & 20.2 &  & 49.3 &  & 48.3 &  & 12.0 &  & 10.5 &  \\ \midrule
Co-training & \multirow{3}{*}{$\{\mathcal{S}, \mathcal{T}_\subset\}$} & 58.4 & {\scriptsize +65.0\%} & 51.3 & {\scriptsize +63.2\%} & 79.3 & {\scriptsize +70.1\%} & 78.1 & {\scriptsize +69.8\%} & 55.7 & {\scriptsize +68.3\%} & 50.3 & {\scriptsize +66.8\%} \\
CutMix~\cite{Wang2022SSDA3DSD} &  & 58.6 & {\scriptsize +65.4\%} & 51.6 & {\scriptsize +63.8\%} & 78.9 & {\scriptsize +69.2\%} & 77.6 & {\scriptsize +68.6\%} & 55.4 & {\scriptsize +67.8\%} & 50.0 & {\scriptsize +66.3\%} \\
SOAP (ours) &  & \textbf{69.4} & \textbf{{\scriptsize +85.5\%}} & \textbf{63.1} & \textbf{{\scriptsize +87.2\%}} & \textbf{84.6} & \textbf{{\scriptsize +82.5\%}} & \textbf{83.5} & \textbf{{\scriptsize +82.4\%}} & \textbf{67.8} & \textbf{{\scriptsize +87.2\%}} & \textbf{63.0} & \textbf{{\scriptsize +88.1\%}} \\ \midrule
SSDA3D~\cite{Wang2022SSDA3DSD} & \multirow{2}{*}{$\{\mathcal{S}, \mathcal{T}_\subset, \mathcal{T}_P\}$} & 67.2 & {\scriptsize +81.4\%} & 59.8 & {\scriptsize +80.5\%} & 84.5 & {\scriptsize +82.2\%} & 83.3 & {\scriptsize +82.0\%} & 65.1 & {\scriptsize +83.0\%} & 59.4 & {\scriptsize +82.0\%} \\
SSDA3D + SOAP (ours) &  & \textbf{70.1} & \textbf{{\scriptsize +86.8\%}} & \textbf{62.6} & \textbf{{\scriptsize +86.2\%}} & \textbf{86.1} & \textbf{{\scriptsize +86.0\%}} & \textbf{84.9} & \textbf{{\scriptsize +85.7\%}} & \textbf{68.5} & \textbf{{\scriptsize +88.3\%}} & \textbf{62.8} & \textbf{{\scriptsize +87.8\%}} \\ \midrule
Oracle & $\{\mathcal{T}\}$ & 77.2 &  & 69.4 &  & 92.1 &  & 91.0 &  & 76.0 &  & 70.1 &  \\ \toprule
\multicolumn{14}{c}{$\mathcal{S}$: labelled source domain; $\mathcal{T}$: labelled target domain; $\mathcal{T}_{\subset}$: small subset of $\mathcal{T}$; $\mathcal{T}_{P}$: pseudo-labelled target domain}
\end{tabular}
}
\vspace{-0.5em}
\caption{Semi-supervised domain adaptation results for nuScenes $\rightarrow$ Waymo, where 1\% of Waymo data is labelled. The percentages represent the amount of the Direct--Oracle domain gap closed.}
\label{tab:ssda3d-nus-waymo}
\end{table*}

\begin{table*}[ht]
\centering
\footnotesize
\setlength{\tabcolsep}{0.4em}
\resizebox{0.8\textwidth}{!}{%
\begin{tabular}{ccllllllllllll}
\toprule
\multirow{2}{*}{\textbf{Method}} & \multirow{2}{*}{\textbf{\begin{tabular}[c]{c}Training\\ Data\end{tabular}}} & \multicolumn{4}{c}{\textbf{Overall}} & \multicolumn{4}{c}{\textbf{0--30\,m}} & \multicolumn{4}{c}{\textbf{30--50\,m}} \\
 &  & \multicolumn{2}{l}{\textbf{mAP}} & \multicolumn{2}{l}{\textbf{NDS}} & \multicolumn{2}{l}{\textbf{mAP}} & \multicolumn{2}{l}{\textbf{NDS}} & \multicolumn{2}{l}{\textbf{mAP}} & \multicolumn{2}{l}{\textbf{NDS}} \\ \midrule
Direct & $\{\mathcal{S}\}$ & 51.7 &  & 69.6 &  & 67.6 &  & 78.9 &  & 27.6 &  & 49.7 &  \\ \midrule
Co-training & \multirow{3}{*}{$\{\mathcal{S}, \mathcal{T}_\subset\}$} & 64.1 & {\scriptsize +38.8\%} & 76.4 & {\scriptsize +36.6\%} & 80.3 & {\scriptsize +50.2\%} & 85.9 & {\scriptsize +46.7\%} & 35.1 & {\scriptsize +21.0\%} & 55.5 & {\scriptsize +23.8\%} \\
CutMix~\cite{Wang2022SSDA3DSD} &  & 63.6 & {\scriptsize +37.2\%} & 76.7 & {\scriptsize +38.2\%} & 82.3 & {\scriptsize +58.1\%} & 87.4 & {\scriptsize +56.7\%} & 29.9 & {\scriptsize +6.4\%} & 52.7 & {\scriptsize +12.3\%} \\
SOAP (ours) &  & \textbf{74.0} & \textbf{{\scriptsize +69.7\%}} & \textbf{83.5} & \textbf{{\scriptsize +74.7\%}} & \textbf{86.4} & \textbf{{\scriptsize +74.3\%}} & \textbf{90.6} & \textbf{{\scriptsize +78.0\%}} & \textbf{49.6} & \textbf{{\scriptsize +61.6\%}} & \textbf{68.0} & \textbf{{\scriptsize +75.0\%}} \\ \midrule
SSDA3D~\cite{Wang2022SSDA3DSD} & \multirow{2}{*}{$\{\mathcal{S}, \mathcal{T}_\subset, \mathcal{T}_P\}$} & 72.7 & {\scriptsize +65.6\%} & 81.3 & {\scriptsize +62.9\%} & 88.1 & {\scriptsize +81.0\%} & \textbf{90.4} & \textbf{{\scriptsize +76.7\%}} & 42.8 & {\scriptsize +42.6\%} & 60.1 & {\scriptsize +42.6\%} \\
SSDA3D + SOAP (ours) &  & \textbf{76.8} & \textbf{{\scriptsize +78.4\%}} & \textbf{83.1} & \textbf{{\scriptsize +72.6\%}} & \textbf{88.3} & \textbf{{\scriptsize +81.8\%}} & \textbf{90.4} & \textbf{{\scriptsize +76.7\%}} & \textbf{52.9} & \textbf{{\scriptsize +70.9\%}} & \textbf{65.6} & \textbf{{\scriptsize +65.2\%}} \\ \midrule
Oracle & $\{\mathcal{T}\}$ & 83.7 &  & 88.2 &  & 92.9 &  & 93.9 &  & 63.3 &  & 74.1 &  \\ \toprule
\multicolumn{14}{c}{$\mathcal{S}$: labelled source domain; $\mathcal{T}$: labelled target domain; $\mathcal{T}_{\subset}$: small subset of $\mathcal{T}$; $\mathcal{T}_{P}$: pseudo-labelled target domain}
\end{tabular}
}
\vspace{-0.5em}
\caption{Semi-supervised domain adaptation results for Waymo $\rightarrow$ nuScenes, where 1\% of nuScenes data is labelled. The percentages represent the amount of the Direct--Oracle domain gap closed.}
\label{tab:ssda3d-waymo-nus}
\end{table*}

For nuScenes evaluation, we consider two primary metrics: mean Average Precision (mAP) and NuScenes Detection Score (NDS). Following the official evaluation, mAP is calculated based on four distance thresholds (0.5, 1.0, 2.0, 4.0) and averaged. As distance-based mAP does not penalize other types of bounding box errors, NDS is used in combination to reflect the average translation, scale, orientation, velocity, and attribute errors for the true positive predictions. All evaluations are performed on the validation split consisting of 150 sequences.

For Waymo evaluation, we use the official evaluation suite and report the Level 1 and Level 2 AP scores. Different from nuScenes mAP, Waymo AP is calculated based on 3D IoU with a threshold of 0.7. Level 1 evaluation includes only objects with more than 5 points within the bounding box, while Level 2 evaluation considers all objects. All evaluations are performed on the validation split consisting of 202 sequences.

\subsection{Unsupervised domain adaptation}
We first evaluate SOAP pseudo-labels in the unsupervised domain adaptation setting, where annotations from the target domain are unavailable. We compare SOAP with two baseline approaches: ``Direct'' and ST3D~\cite{yang2021st3d}. Direct is where a few-frame detector is trained on the source domain and directly evaluated on the target domain data.
ST3D is a SOTA unsupervised domain adaptation method based on pseudo-labelling and self-training. 

In the baseline comparison, ST3D utilizes the Direct model to generate pseudo-labels for self-training. To demonstrate the quality of the SOAP pseudo-labels and the complementary nature of SOAP with other approaches, we further consider ST3D + SOAP, where ST3D uses SOAP pseudo-labels. Both ST3D experiments use the official code release.

SOAP is validated using two object detection architectures: CenterPoint\cite{yin2021centerpoint} and VoxelNeXt~\cite{chen2023voxelnext}. CenterPoint is a widely-adopted voxel-based dense 3D object detector. VoxelNeXt is a SOTA architecture representing recent advances in fully-sparse 3D object detectors. Both architectures are based on the implementation in the open-source library OpenPCDet. We use the Direct model predictions as few-frame predictions to construct final SOAP pseudo-labels. More implementation detail and hyper-parameters can be found in the supplementary material. The main results are shown in \Cref{tab:st3d-nus-waymo,tab:st3d-waymo-nus}.

\paragraph{Pseudo-label performance:}
Overall, SOAP pseudo-labels improve over the Direct pseudo-labelling baseline by a significant margin. In the nuScenes $\rightarrow$ Waymo setting (\cref{tab:st3d-nus-waymo}), both architectures receive a 25--30 point improvement in mAP, with over 50\% domain gap closed. SOAP is as effective in the Waymo $\rightarrow$ nuScenes setting (\cref{tab:st3d-waymo-nus}), with both architectures showing a 10-point improvement in mAP and over 30\% domain gap closed.

More importantly, SOAP consistently improves object pseudo-labels at different ranges, with the largest improvements observed for objects at the 30--50\,m range in all settings, closing 40--60\% of the domain gap. We suppose this is because objects farther away from the sensor have sparse point clouds and are sometimes occluded. The results highlight the benefits of full-sequence aggregation, as far objects are densified, and occlusion can be alleviated by aggregating multiple viewpoints. The supplementary material includes qualitative examples that further illustrate the accuracy of SOAP pseudo-labels at long range.

\paragraph{Adaptation performance:}
\begin{table*}[ht]
\centering
\footnotesize
\setlength{\tabcolsep}{0.4em}
\resizebox{0.7\textwidth}{!}{%
\begin{tabular}{c|ccccc|llllllll}
\toprule
\multirow{2}{*}{\textbf{}} & \multirow{2}{*}{\textbf{1-frame}} & \multirow{2}{*}{\textbf{5-frame}} & \multirow{2}{*}{\textbf{SFA}} & \multirow{2}{*}{\textbf{QST}} & \multirow{2}{*}{\textbf{SCP}} & \multicolumn{4}{c}{\textbf{Overall}} & \multicolumn{4}{c}{\textbf{Stationary}} \\
 &  &  &  &  &  & \multicolumn{2}{l}{\textbf{Level 1}} & \multicolumn{2}{l}{\textbf{Level 2}} & \multicolumn{2}{l}{\textbf{Level 1}} & \multicolumn{2}{l}{\textbf{Level 2}} \\ \midrule
\textbf{(a)} & \textbf{\checkmark} &  &  &  &  & 7.0 & {\scriptsize -23.5\%} & 6.0 & {\scriptsize -22.0\%} & 5.9 & {\scriptsize -21.9\%} & 4.9 & {\scriptsize -20.5\%} \\
\textbf{(b)} &  & \textbf{\checkmark} &  &  &  & 20.4 &  & 17.5 &  & 18.6 &  & 15.6 &  \\ \midrule
\textbf{(c)} &  & \textbf{\checkmark} & \textbf{\checkmark} &  &  & 35.7 & {\scriptsize +36.6\%} & 31.2 & {\scriptsize +35.6\%} & 38.0 & {\scriptsize +50.1\%} & 32.6 & {\scriptsize +48.3\%} \\
\textbf{(d)} &  & \textbf{\checkmark} & \textbf{\checkmark} & \textbf{\checkmark} &  & 46.7 & {\scriptsize +46.1\%} & 41.6 & {\scriptsize +46.2\%} & 52.1 & {\scriptsize +57.7\%} & 45.8 & {\scriptsize +57.9\%} \\
\textbf{(e)} &  & \textbf{\checkmark} & \textbf{\checkmark} & \textbf{\checkmark} & \textbf{\checkmark} & \textbf{50.9} & \textbf{{\scriptsize +53.4\%}} & \textbf{45.6} & \textbf{{\scriptsize +53.8\%}} & \textbf{57.2} & \textbf{{\scriptsize +66.4\%}} & \textbf{50.6} & \textbf{{\scriptsize +67.0\%}} \\ \midrule
\textbf{Oracle} &  &  &  &  &  & 77.5 &  & 69.7 &  & 76.7 &  & 67.8 &  \\ \bottomrule
\end{tabular}
}
\vspace{-0.5em}
\caption{Ablation study results for nuScenes $\rightarrow$ Waymo unsupervised domain adaptation.  The percentages represent the amount of domain gap closed relative to the 5-frame baseline detector.}
\vspace{-1em}
\label{tab:ablation}
\end{table*}

SOAP complements the SOTA domain adaptation technique ST3D. We observe that when equipped with SOAP pseudo-labels, ST3D + SOAP provides better overall performance than ST3D. In the nuScenes $\rightarrow$ Waymo setting (\cref{tab:st3d-nus-waymo}), ST3D + SOAP closes 20--30\% more domain gap than ST3D. While the difference is smaller in the Waymo $\rightarrow$ nuScenes setting (\cref{tab:st3d-waymo-nus}), there is still a noticeable improvement, with around 5\% more domain gap closed for mAP and over 10\% more domain gap closed for NDS. Moreover, the aforementioned improvement over far objects can also be seen after self-training with ST3D. Using SOAP pseudo-labels achieves significantly higher performance compared to ST3D.

\subsection{Semi-supervised domain adaptation}
In the semi-supervised domain adaptation setting, where a small amount of target domain annotations are available for training, SOAP can also be used to improve pseudo-label quality. To demonstrate this, we compare SOAP with three methods: Direct, Co-training\cite{Wang2022SSDA3DSD}, and SSDA3D~\cite{Wang2022SSDA3DSD}. As in the unsupervised case, Direct is where the model is trained only on source domain data. Co-training is where the model is trained with a combination of labelled source and labelled target domain data. SSDA3D is a recent SOTA semi-supervised domain adaptation technique. SSDA3D consists of a pseudo-labelling stage with inter-domain CutMix augmentation to improve pseudo-label quality (which we denote CutMix), followed by a target domain training stage with intra-domain MixUp augmentation as regularization. Additionally, we explore the SSDA3D + SOAP configuration, where we replace the SSDA3D pseudo-labels with SOAP pseudo-labels for second-stage target domain training. 

Following SSDA3D~\cite{Wang2022SSDA3DSD}, we use CenterPoint for experiments in this section and consider 1\% sequences labelled in the target domain. Note that, unlike experiments in SSDA3D, we uniformly sample entire sequences instead of individual frames. Following how SSDA3D's pseudo-labelling model is trained, the SOAP model is trained on both labelled source and labelled target sequences with CutMix, among other standard augmentation, applied to aggregated point clouds. We use the SSDA3D CutMix predictions as sparse predictions to construct final SOAP pseudo-labels.

The main results are shown in \Cref{tab:ssda3d-nus-waymo,tab:ssda3d-waymo-nus}.

\paragraph{Pseudo-label performance:}
Compared to the pseudo-labels generated by Co-training and SSDA3D CutMix, SOAP pseudo-labels are much more accurate, closing 85.5\% and 69.7\% domain gap for nuScenes $\rightarrow$ Waymo (\cref{tab:ssda3d-nus-waymo}) and Waymo $\rightarrow$ nuScenes (\cref{tab:ssda3d-waymo-nus}), respectively. Similar to the results in the unsupervised settings, the improvement is particularly noticeable for objects at 30--50\,m, further illustrating the benefits of full-sequence aggregation.

\paragraph{Adaptation performance:}
SOAP improves the already impressive domain adaptation performance achieved by SSDA3D, closing 5--10\% more overall performance gap in both settings. Moreover, we observe the improvements in pseudo-labels for objects at 30--50\,m translate to the model after adaptation. In the Waymo $\rightarrow$ nuScenes setting (\cref{tab:ssda3d-waymo-nus}), training with SOAP pseudo-labels achieves 10.1\% higher mAP and 5.5\% higher NDS for 30--50\,m objects.

\subsection{Ablation study}
In this section, we investigate the benefits of QST and SCP in the nuScenes $\rightarrow$ Waymo unsupervised domain adaptation setting, using the VoxelNeXt architecture. The results are presented in \Cref{tab:ablation}.

We use the 1-frame and 5-frame detectors as baselines in lines (a) and (b), respectively, and progressively introduce each component of SOAP. In line (c), SFA augments the 5-frame detector with stationary object predictions using aggregated point clouds. The model is trained by naively filtering object speed based on a threshold of 0.2\,m/s. Line (d) enables QST, replacing naive filtering. While the SFA pseudo-labels improve over both the 1-frame and 5-frame baselines, especially for stationary objects, it is still significantly outperformed by QST. This highlights both the effectiveness of full-sequence aggregation in cross-sensor settings and the importance of constructing robust training labels for stationary objects using QST. Moreover, incorporating SCP in line (e) further improves the AP by over 4\%, demonstrating the benefit of exploiting the stationarity of the detected objects.

\section{Limitations} \label{sec:limitations}
SOAP has three principal limitations. First, constructing aggregated point clouds requires the point cloud data to be collected sequentially and the ego pose estimates to be available. This is applicable in most current self-driving datasets but may not work in other applications where sequential information is not available. Second, SOAP assumes the ego vehicle--hence the sensor--is moving relative to the static environment. It is not applicable to roadside detection where the sensor stays stationary. Finally, since SOAP is designed to detect stationary objects to augment sparse pseudo-labels, for objects like pedestrians or environments with mostly dynamic objects, SOAP may be less effective. However, in major realistic self-driving datasets~\cite{caesar2020nuscenes,sun2020waymo,wilson2021argoverse,kesten2019lyft}, we find that at least two thirds of vehicles are stationary at some point in the sequence, making SOAP effective for practical applications. Detailed statistics can be found in the supplementary material.

\section{Conclusion} \label{sec:conclusion}
We have presented Stationary Object Aggregation Pseudo-labelling (SOAP), a novel method that utilizes full-sequence scene-level aggregation to generate high-quality pseudo-labels for the cross-sensor domain adaptation setting. We have provided extensive evaluation that demonstrates SOAP can provide high-quality pseudo-labels and improves the already impressive results achieved by SOTA methods such as ST3D and SSDA3D.

As future work,  we want to exploring the benefits of tracking and second-stage refinement, as used by in-domain pseudo-labelling methods, in the domain adaptation setting. It will also be interesting to explore the synergy of SOAP with other domain adaptation approaches. 

 \clearpage

\begin{appendix}
\begin{strip}
 \centering
 \Large \textbf{Supplementary Material: Cross-sensor Domain Adaptation for 3D Object Detection Using Stationary Object Aggregation Pseudo-labelling}
 \vspace{1em}
\end{strip}

\section{Implementation Details}
In this section, we include the implementation details for the experiments presented in the main text.

\subsection{Point Cloud Input}
\paragraph{Single- and few-frame input:}
As detailed in the main text, the nuScenes~\cite{caesar2020nuscenes} and Waymo~\cite{sun2020waymo} datasets have different sensor ranges and point cloud features. In our experiments, we match the input point cloud format. Specifically, the input point cloud range is $[-75, 75]$\,m for both $x$ and $y$ dimensions, and $[-2, 4]$\,m for the $z$ dimension. For few-frame models, we use five features $(x, y, z, i, e, t)$ for each point, where $(x,y,z)$ are the point location, $i$ is the intensity normalized to $[0, 1]$, $e$ is the elongation, and $t$ is the timestamp offset in seconds. For the single-frame model, we exclude timestamp offset $t$. 

Since the nuScenes dataset does not provide elongation information, we set $e=0$ for all nuScenes point clouds. Moreover, following previous work~\cite{yang2021st3d,Wang2022SSDA3DSD}, we apply a $+1.8$\,m offset to the $z$ dimension to approximately transform the nuScenes point clouds from sensor frame to ego vehicle frame.

\paragraph{Full-sequence input:} For full-sequence models, we only use the $(x,y,z)$ channels. To reduce training time and memory consumption, we pre-compute the aggregated point cloud for each sequence and perform a voxel-downsampling step with $3.25\,\mathrm{cm}^3$ voxels. During training, the pre-computed aggregated point clouds are transformed using pose transformations and further uniformly downsampled to at most $1{,}000{,}000$ points.

Similar to single and few-frame input, a $+1.8\,\mathrm{m}$ offset is applied to nuScenes aggregated point clouds.

\subsection{Architecture}
We use CenterPoint~\cite{yin2021centerpoint} and VoxelNeXt~\cite{chen2023voxelnext} implemented in the open-source framework OpenPCDet\footnote{https://github.com/open-mmlab/OpenPCDet} with minor modifications to make the models compatible to both nuScenes and Waymo datasets.

\paragraph{Voxelization:} The point cloud is voxelized using a voxel size of $(7.5\,\mathrm{cm},7.5\,\mathrm{cm},15\,\mathrm{cm})$. For each point cloud, we use at most $500{,}000$ voxels, with each voxel containing at most $10$ points. 

\paragraph{Backbone:} We adopt the backbone used in nuScenes models for both datasets. Detailed configurations can be found in the OpenPCDet repository.

\paragraph{Detection heads:} Since our models are trained for only Vehicle\,/\,Car class, We use a single detection head for both architectures. For few-frame models, an additional head with $2$ convolution layers is added to regress the velocity $(v_x,v_y)$.

\subsection{Training}
All models are trained with a total batch size of 32, over multiple GPUs. We use the Adam optimizer with a one cycle learning rate schedule. 

\paragraph{Baseline:} All single-frame and few-frame models are trained for 36 epochs. The learning rate is set to $0.001$ for nuScenes models and $0.003$ for Waymo models.

\paragraph{ST3D~\cite{yang2021st3d}} For the nuScenes $\rightarrow$ Waymo direction, the models are fine-tuned for 12 epochs with a learning rate of $0.0001$. The positive and negative confidence thresholds for pseudo-labelling are set to $(0.5,0.3)$ for Direct pseudo-labels and $(0.1,0.05)$ for SOAP pseudo-labels. For the Waymo $\rightarrow$ nuScenes direction, the models are fine-tuned for 6 epochs with a learning rate of $0.0003$. The positive and negative confidence thresholds are set to $(0.6,0.2)$ for Direct pseudo-labels and $(0.3,0.2)$ for SOAP pseudo-labels. In all experiments, Direct pseudo-labels are updated every 2 epochs using the memory ensemble proposed in ST3D.

\paragraph{SSDA3D~\cite{Wang2022SSDA3DSD}} Both stages in SSDA3D experiments follow the baseline training configurations. The CutMix and MixUp augmentation probabilities are set to $0.5$. Predictions from the first stage models are filtered by a confidence threshold of $0.3$ to construct the corresponding pseudo-labels for second stage training. When SOAP predictions are used, confidence thresholds of $0.15$ and $0.25$ are used to construct Waymo and nuScenes pseudo-labels, respectively.

\paragraph{SOAP} The SOAP model is initialized with the weights from a corresponding few-frame model and trained for an additional 12 epochs with a learning rate of $0.001$ for nuScenes and $0.003$ for Waymo. As described in the main text, the annotations are constructed using QST. We set the QSS threshold $\epsilon$ to $0.7$ and $0.85$ for nuScenes and Waymo datasets, respectively.

\subsection{Post-processing}
The post-processing for each dataset follows the implementation in OpenPCDet.

\paragraph{nuScenes:} The predictions are filtered with a confidence threshold of $0.1$ and a range of $[-61.2, 61.2]$\,m for both $x$ and $y$, and $[-10,10]$\,m for $z$. NMS is performed on the best $1000$ predictions using an IoU threshold of $0.2$, with at most $83$ predictions retained.

\paragraph{Waymo:} The predictions are filtered with a confidence threshold of $0.1$ and a range of $[-75.2, 75.2]$\,m for both $x$ and $y$, and $[-2,4]$\,m for $z$. NMS is performed on the best $4096$ predictions using an IoU threshold of $0.7$, with at most $500$ predictions retained.

\begin{figure}
    \centering
    \includegraphics[width=0.8\linewidth]{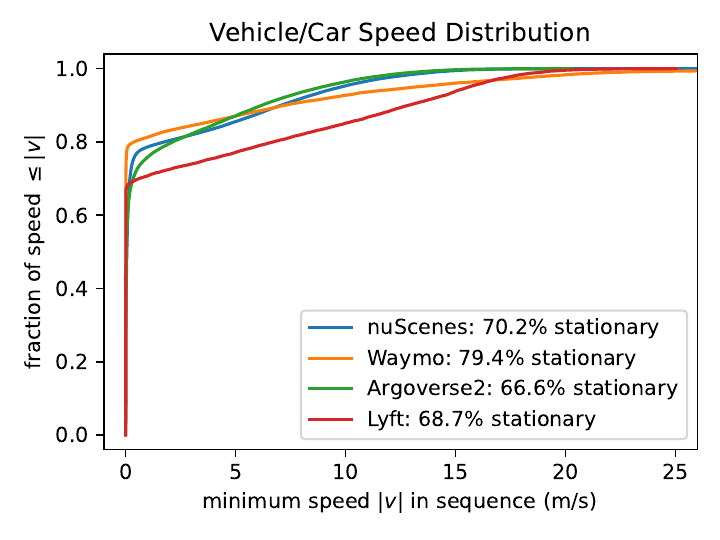}
    \caption{Cumulative distribution for Vehicle\,/\,Car speed in realistic self-driving datasets.}
    \label{fig:speed-cdf}
\end{figure}

\begin{figure}
    \centering
    \includegraphics[width=0.8\linewidth]{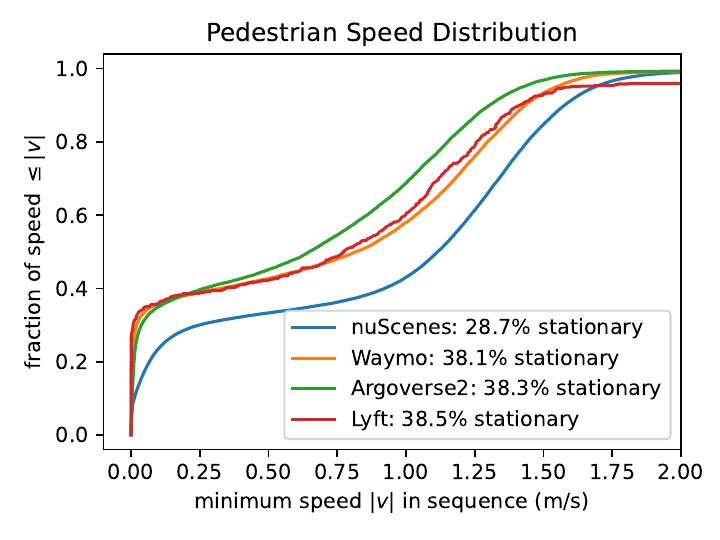}
    \caption{Cumulative distribution for Pedestrian speed in realistic self-driving datasets.}
    \label{fig:speed-cdf-ped}
\end{figure}

\begin{figure}
    \centering
    \includegraphics[width=0.8\linewidth]{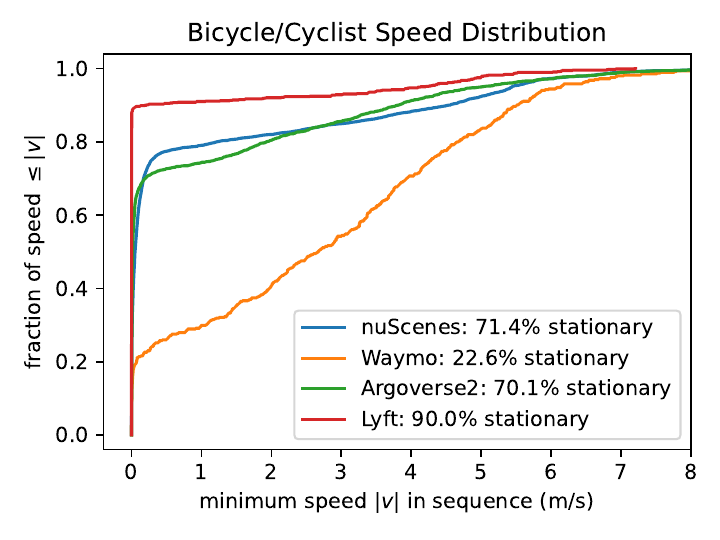}
    \caption{Cumulative distribution for Bicycle\,/\,Cyclist speed in realistic self-driving datasets.}
    \label{fig:speed-cdf-cyc}
\end{figure}

\begin{table*}[]
\centering
\footnotesize
\setlength{\tabcolsep}{0.35em}
\begin{tabular}{ccllllllllllllllll}
\toprule
\multirow{2}{*}{\textbf{Method}} & \multirow{2}{*}{\textbf{\begin{tabular}[c]{@{}c@{}}Training\\ Data\end{tabular}}} & \multicolumn{4}{c}{\textbf{Stationary ($<$0.2 m/s)}} & \multicolumn{4}{c}{\textbf{Slow (0.2-1 m/s)}} & \multicolumn{4}{c}{\textbf{Medium (1-3 m/s)}} & \multicolumn{4}{c}{\textbf{Fast (3-10 m/s)}} \\
 &  & \multicolumn{2}{l}{\textbf{Level 1}} & \multicolumn{2}{l}{\textbf{Level 2}} & \multicolumn{2}{l}{\textbf{Level 1}} & \multicolumn{2}{l}{\textbf{Level 2}} & \multicolumn{2}{l}{\textbf{Level 1}} & \multicolumn{2}{l}{\textbf{Level 2}} & \multicolumn{2}{l}{\textbf{Level 1}} & \multicolumn{2}{l}{\textbf{Level 2}} \\ \midrule
Direct & \multirow{2}{*}{$\{\mathcal{S}\}$} & 21.5 &  & 18.0 &  & \textbf{24.3} &  & \textbf{22.3} &  & \textbf{27.5} &  & \textbf{25.5} &  & \textbf{27.1} &  & \textbf{25.4} &  \\
 SOAP (ours) &  & \textbf{56.2} & \textbf{{\tiny +63.3\%}} & \textbf{49.5} & \textbf{{\tiny +63.5\%}} & 23.9 & {\tiny -0.9\%} & 22.0 & {\tiny -0.7\%} & 24.4 & {\tiny -7.4\%} & 22.6 & {\tiny -7.4\%} & 25.7 & {\tiny -2.8\%} & 24.1 & {\tiny -2.7\%} \\ \midrule
 ST3D~\cite{yang2021st3d} & \multirow{2}{*}{$\{\mathcal{S},\mathcal{T}_P\}$} & 27.1 & {\tiny +10.2\%} & 22.8 & {\tiny +9.7\%} & 26.1 & {\tiny +3.9\%} & 24.0 & {\tiny +4.0\%} & 23.9 & {\tiny -8.6\%} & 22.1 & {\tiny -8.6\%} & 28.8 & {\tiny +3.4\%} & 27.0 & {\tiny +3.4\%} \\
 ST3D + SOAP (ours) &  & \textbf{48.6} & \textbf{{\tiny +49.5\%}} & \textbf{41.7} & \textbf{{\tiny +47.8\%}} & \textbf{32.6} & \textbf{{\tiny +17.9\%}} & \textbf{30.0} & \textbf{{\tiny +17.9\%}} & \textbf{32.2} & \textbf{{\tiny +11.2\%}} & \textbf{29.9} & \textbf{{\tiny +11.2\%}} & \textbf{34.4} & \textbf{{\tiny +14.7\%}} & \textbf{32.3} & \textbf{{\tiny +14.5\%}} \\ \midrule 
 Oracle & $\{\mathcal{T}\}$ & 76.3 &  & 67.6 &  & 70.6 &  & 65.3 &  & 69.6 &  & 64.9 &  & 76.7 &  & 73.1 &  \\ \toprule
\multicolumn{18}{c}{$\mathcal{S}$: labelled source domain; $\mathcal{T}$: labelled target domain; $\mathcal{T}_{\subset}$: small subset of $\mathcal{T}$; $\mathcal{T}_{P}$: pseudo-labelled target domain}
\end{tabular}
\caption{Unsupervised domain adaptation results for nuScenes $\rightarrow$ Waymo, where Waymo dataset is unlabelled, split based on object speed. The percentages represent the amount of the Direct–Oracle domain gap closed.}
\label{tab:st3d-nus-waymo-speed}
\end{table*}
\begin{table*}[]
\centering
\footnotesize
\setlength{\tabcolsep}{0.4em}
\begin{tabular}{ccllllllllllllllll}
\toprule
\multirow{2}{*}{\textbf{Method}} & \multirow{2}{*}{\textbf{\begin{tabular}[c]{@{}c@{}}Training\\ Data\end{tabular}}} & \multicolumn{4}{c}{\textbf{Stationary ($<$0.2 m/s)}} & \multicolumn{4}{c}{\textbf{Slow (0.2-1 m/s)}} & \multicolumn{4}{c}{\textbf{Medium (1-3 m/s)}} & \multicolumn{4}{c}{\textbf{Fast (3-10 m/s)}} \\
 &  & \multicolumn{2}{l}{\textbf{Level 1}} & \multicolumn{2}{l}{\textbf{Level 2}} & \multicolumn{2}{l}{\textbf{Level 1}} & \multicolumn{2}{l}{\textbf{Level 2}} & \multicolumn{2}{l}{\textbf{Level 1}} & \multicolumn{2}{l}{\textbf{Level 2}} & \multicolumn{2}{l}{\textbf{Level 1}} & \multicolumn{2}{l}{\textbf{Level 2}} \\ \midrule
Direct & $\{\mathcal{S}\}$ & 21.5 &  & 18.0 &  & 24.3 &  & 22.3 &  & 27.5 &  & 25.5 &  & 27.1 &  & 25.4 &  \\ \midrule
Co-training & \multirow{3}{*}{$\{\mathcal{S},\mathcal{T}_{\subset}\}$} & 58.1 & {\tiny +66.2\%} & 50.0 & {\tiny +64.4\%} & 50.1 & {\tiny +52.3\%} & 46.2 & {\tiny +52.1\%} & 44.5 & {\tiny +37.7\%} & 41.3 & {\tiny +37.6\%} & 51.5 & {\tiny +47.9\%} & 48.7 & {\tiny +47.5\%} \\
CutMix~\cite{Wang2022SSDA3DSD} &  & 57.4 & {\tiny +65.1\%} & 49.6 & {\tiny +63.4\%} & \textbf{51.1} & \textbf{{\tiny +54.3\%}} & \textbf{47.1} & \textbf{{\tiny +54.1\%}} & \textbf{48.0} & \textbf{{\tiny +45.4\%}} & \textbf{44.5} & \textbf{{\tiny +45.2\%}} & \textbf{57.2} & \textbf{{\tiny +59.2\%}} & \textbf{54.2} & \textbf{{\tiny +58.7\%}} \\
SOAP (ours) &  & \textbf{71.9} & \textbf{{\tiny +91.3\%}} & \textbf{64.6} & \textbf{{\tiny +93.6\%}} & 39.4 & {\tiny +30.6\%} & 36.4 & {\tiny +30.8\%} & 38.6 & {\tiny +24.6\%} & 35.8 & {\tiny +24.5\%} & 51.8 & {\tiny +48.5\%} & 49.0 & {\tiny +48.2\%} \\ \midrule
SSDA3D~\cite{Wang2022SSDA3DSD} & \multirow{2}{*}{$\{\mathcal{S},\mathcal{T}_{\subset},\mathcal{T}_P\}$} & 66.5 & {\tiny +81.6\%} & 58.2 & {\tiny +80.7\%} & 56.3 & {\tiny +64.8\%} & 52.0 & {\tiny +64.6\%} & 53.8 & {\tiny +58.2\%} & 50.0 & {\tiny +58.2\%} & 62.6 & {\tiny +69.7\%} & 59.4 & {\tiny +69.2\%} \\
SSDA3D + SOAP (ours) &  & \textbf{70.1} & \textbf{{\tiny +88.0\%}} & \textbf{61.6} & \textbf{{\tiny +87.5\%}} & \textbf{58.7} & \textbf{{\tiny +69.7\%}} & \textbf{54.3} & \textbf{{\tiny +69.6\%}} & \textbf{54.6} & \textbf{{\tiny +60.0\%}} & \textbf{50.7} & \textbf{{\tiny +59.9\%}} & \textbf{63.6} & \textbf{{\tiny 71.8\%}} & \textbf{60.3} & \textbf{{\tiny +71.2\%}} \\ \midrule
Oracle & $\{\mathcal{T}\}$ & 76.7 &  & 67.8 &  & 73.7 &  & 68.2 &  & 72.6 &  & 67.6 &  & 78.0 &  & 74.4 &  \\ \midrule
\multicolumn{18}{c}{$\mathcal{S}$: labelled source domain; $\mathcal{T}$: labelled target domain; $\mathcal{T}_{\subset}$: small subset of $\mathcal{T}$; $\mathcal{T}_{P}$: pseudo-labelled target domain}
\end{tabular}
\caption{Semi-supervised domain adaptation results for nuScenes $\rightarrow$ Waymo, where 1\% of Waymo data is labelled, split based on object speed. The percentages represent the amount of the Direct--Oracle domain gap closed.}
\label{tab:ssda3d-nus-waymo-speed}
\end{table*}

\paragraph{SCP} The SOAP preditions undergo the SCP step, which clusters and filters predictions in the global coordinate system. The cluster size threshold $\eta$ depends on the frame rate of the dataset, so we use $\eta=10$ for Waymo (10\,Hz) and $\eta=2$ for nuScenes (2\,Hz). The cluster threshold $\mu$ for both SCP and WBF are set to $0.5$.

\section{Speed Statistics in Self-driving Datasets}

As mentioned in the main text, we observe that stationary objects are a statistically important component of object detection. In~\cref{fig:speed-cdf}, we present the cumulative distribution of speeds for the Vehicle\,/\,Car class in four realistic self-driving datasets: nuScenes~\cite{caesar2020nuscenes}, Waymo~\cite{sun2020waymo}, Lyft~\cite{kesten2019lyft}, and Argoverse2~\cite{wilson2021argoverse}. In all  datasets, we observe a significant proportion of objects are stationary ($|v| < 0.2\,\mathrm{m/s}$) at some point in the sequence, ranging from 66.6\% in Argoverse2 to 79.4\% in Waymo. 

The corresponding distribution for pedestrian and bicycle/cyclist are shown in \cref{fig:speed-cdf-ped,fig:speed-cdf-cyc}. For bicycle/cyclist class, we observe similar statistics in nuScenes, Argoverse and Lyft dataset. Note that in the Waymo dataset, only bicycles with riders are labelled, hence the much lower percentage compared to other datasets.

For pedestrian, however, the percentage of objects that are stationary at some point in the sequence is significantly smaller. As mentioned in the limitation section in the main text, this may limit the effectiveness of our approach to these classes.

\section{Additional Results}
We include additional evaluation based on speed for nuScenes $\rightarrow$ Waymo CenterPoint models in \Cref{tab:st3d-nus-waymo-speed,tab:ssda3d-nus-waymo-speed}. First, we notice that in all cases, the stationary performance of SOAP pseudo-labels and models fine-tuned with SOAP pseudo-labels exceeds SOTA by a significant margin, highlighting the effectiveness of our proposed method. Second, interestingly, while the pseudo-label performance for dynamic objects is on par or worse than the few-frame baseline (Direct and Co-training), after fine-tuning with the SOAP pseudo-labels using ST3D or SSDA3D, the dynamic performance is consistently better than SOTA methods.

\section{Qualitative Results}
\begin{figure*}[ht]
    \centering
    \begin{subfigure}[b]{0.32\textwidth}
        \includegraphics[width=\textwidth]{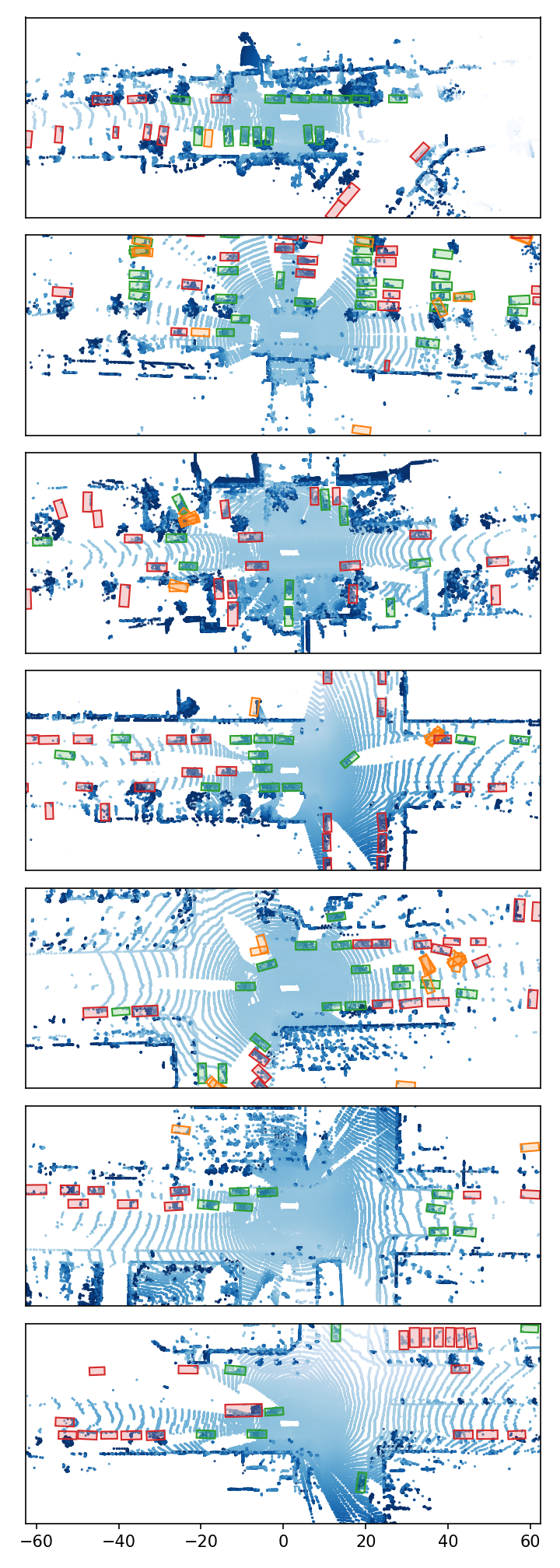}
        \caption{Direct}
    \end{subfigure}
    \begin{subfigure}[b]{0.32\textwidth}
        \includegraphics[width=\textwidth]{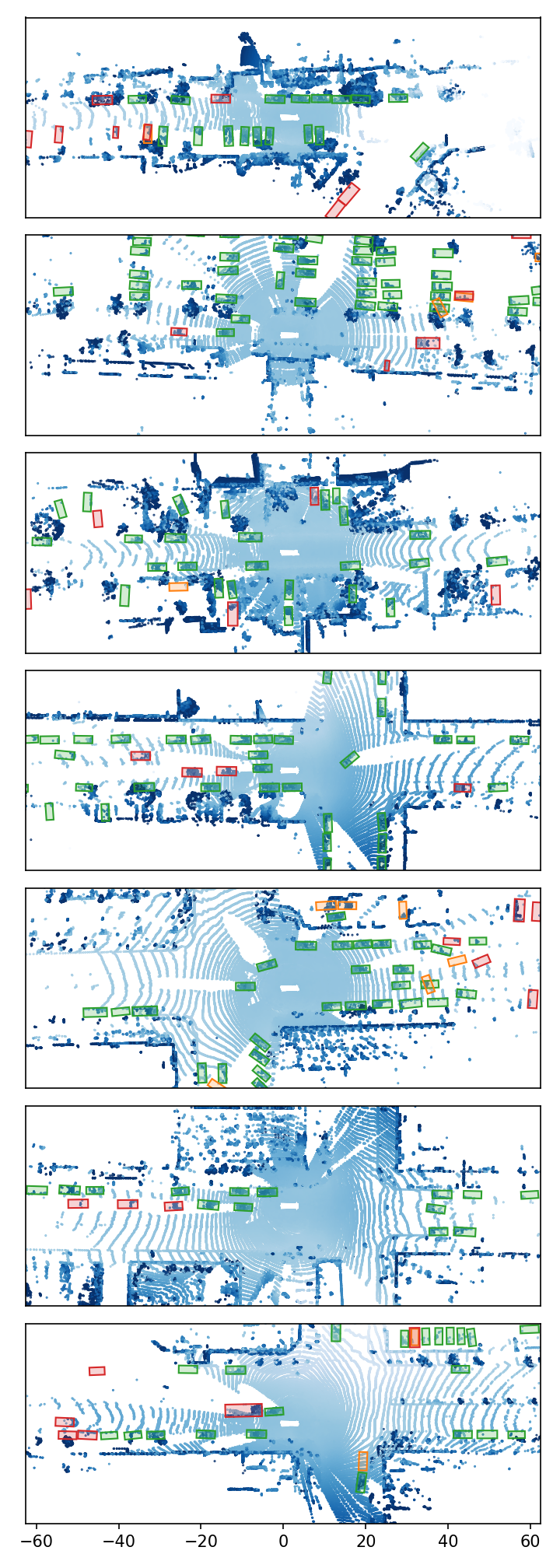}
        \caption{SOAP}
    \end{subfigure}
    \begin{subfigure}[b]{0.32\textwidth}
        \includegraphics[width=\textwidth]{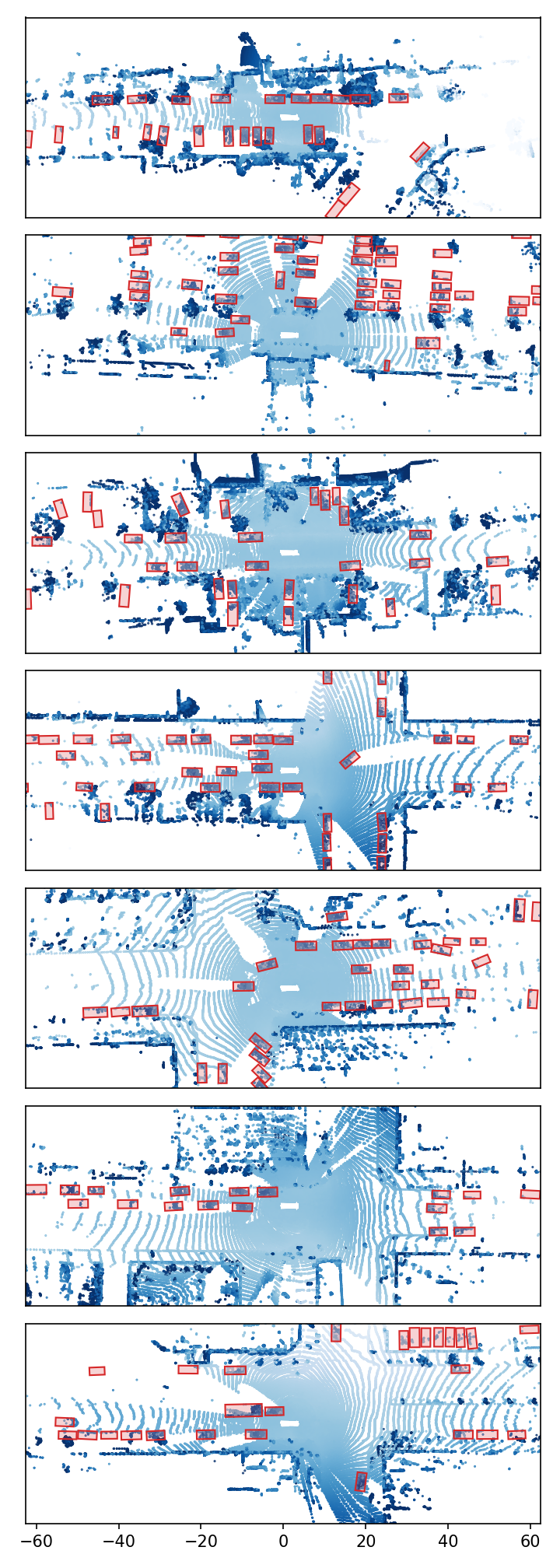}
        \caption{Ground Truth}
    \end{subfigure}
    \caption{Examples of pseudo-labels generated by different methods in nuScenes $\rightarrow$ Waymo unsupervised domain adaptation setting, and the corresponding ground truth labels. Green represents true positive pseudo-labels, orange represents false positive pseudo-labels, and red represents false negative pseudo-labels.}
    \label{fig:qualitative-uda-nus-waymo}
\end{figure*}

\begin{sidewaysfigure*}[ht]
    \centering
    \begin{subfigure}[b]{0.24\textwidth}
        \includegraphics[width=\textwidth]{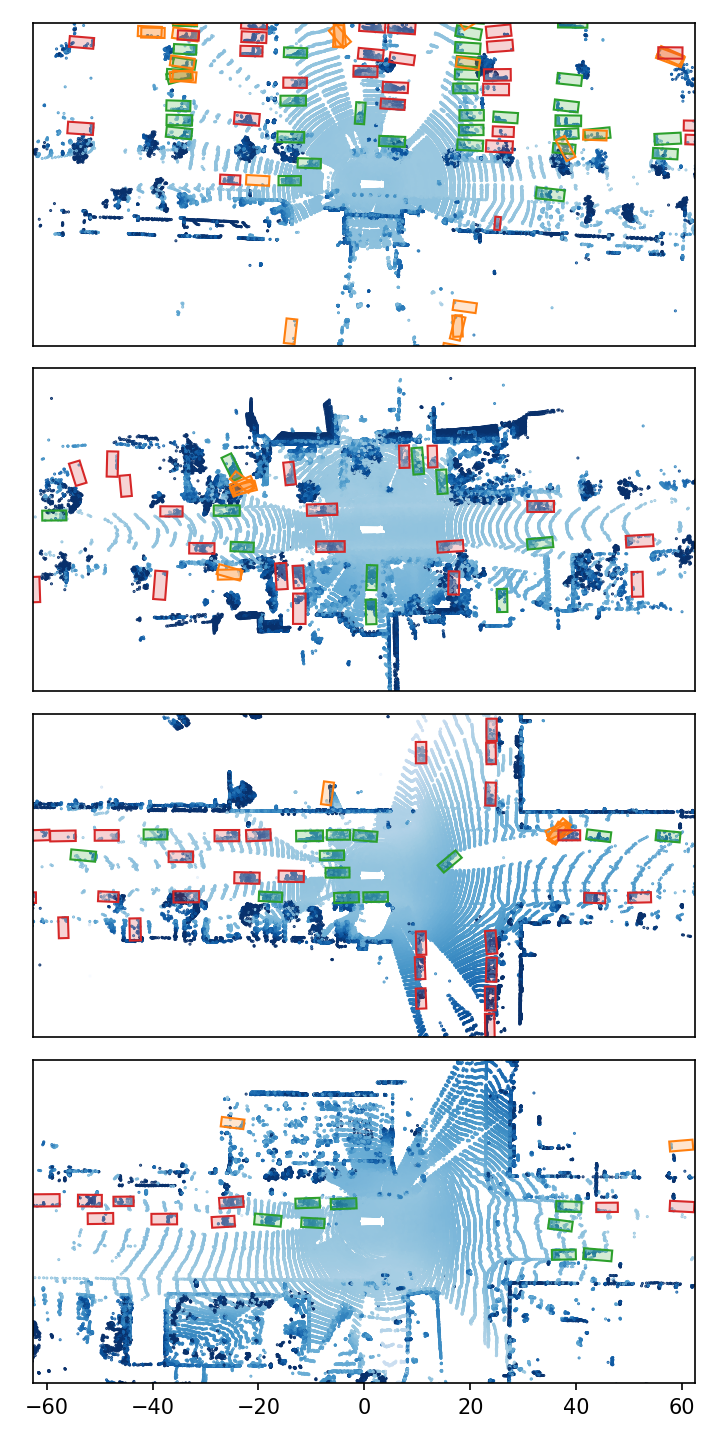}
        \caption{Direct}
    \end{subfigure}
    \begin{subfigure}[b]{0.24\textwidth}
        \includegraphics[width=\textwidth]{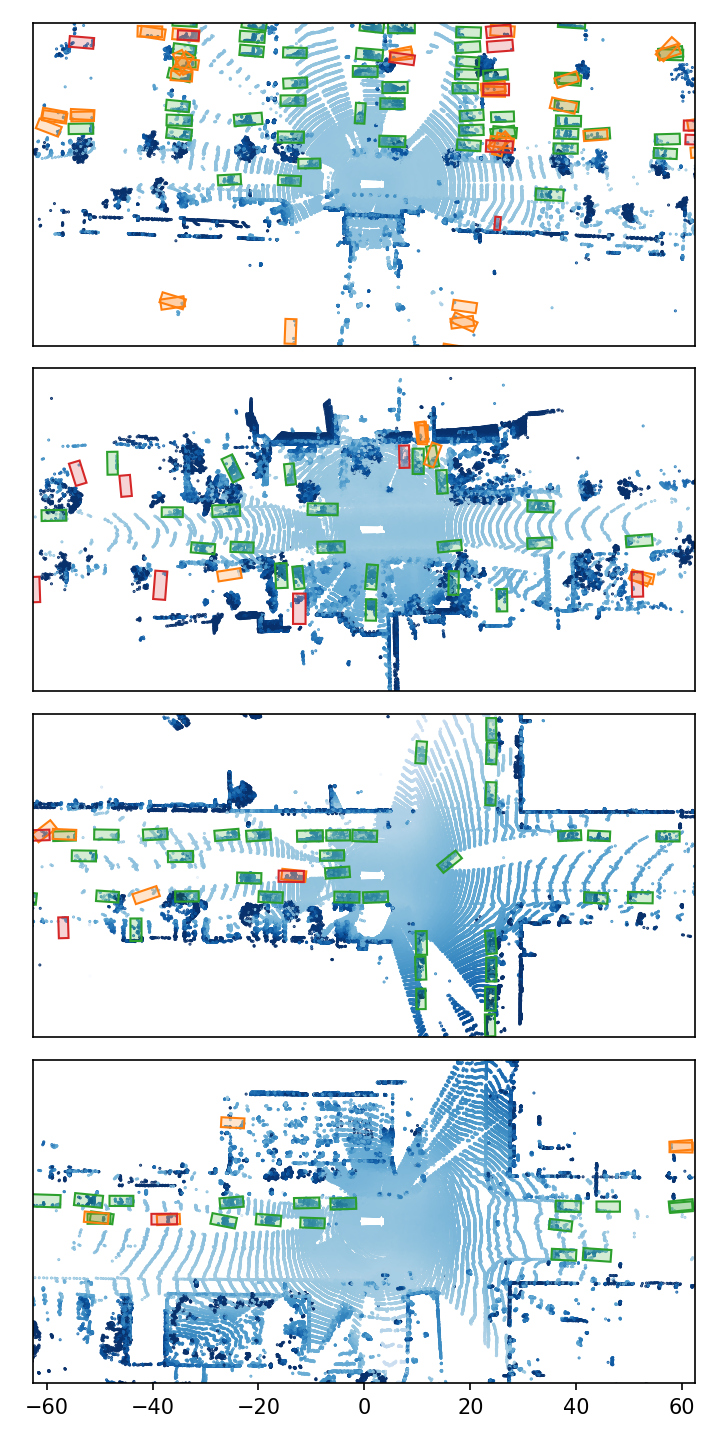}
        \caption{Co-training}
    \end{subfigure}
    \begin{subfigure}[b]{0.24\textwidth}
        \includegraphics[width=\textwidth]{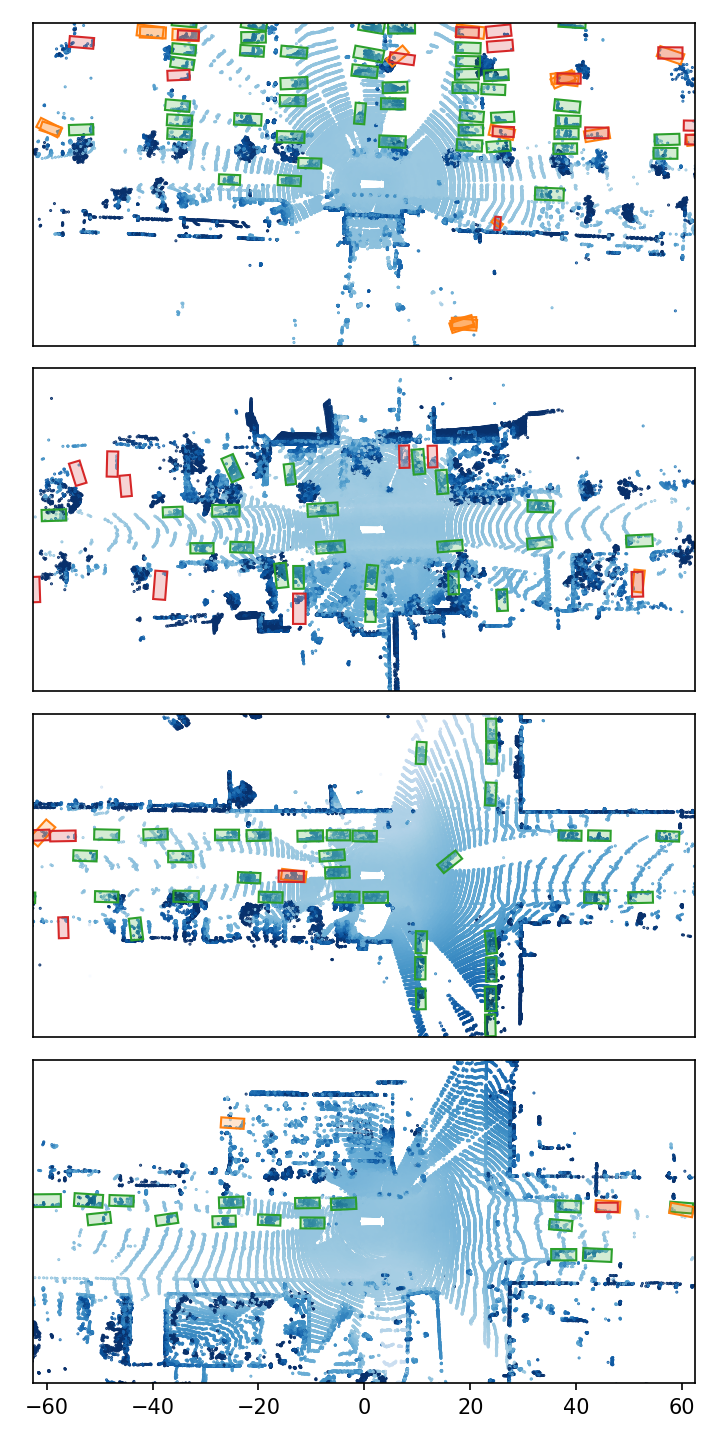}
        \caption{CutMix (SSDA3D)}
    \end{subfigure}
    \begin{subfigure}[b]{0.24\textwidth}
        \includegraphics[width=\textwidth]{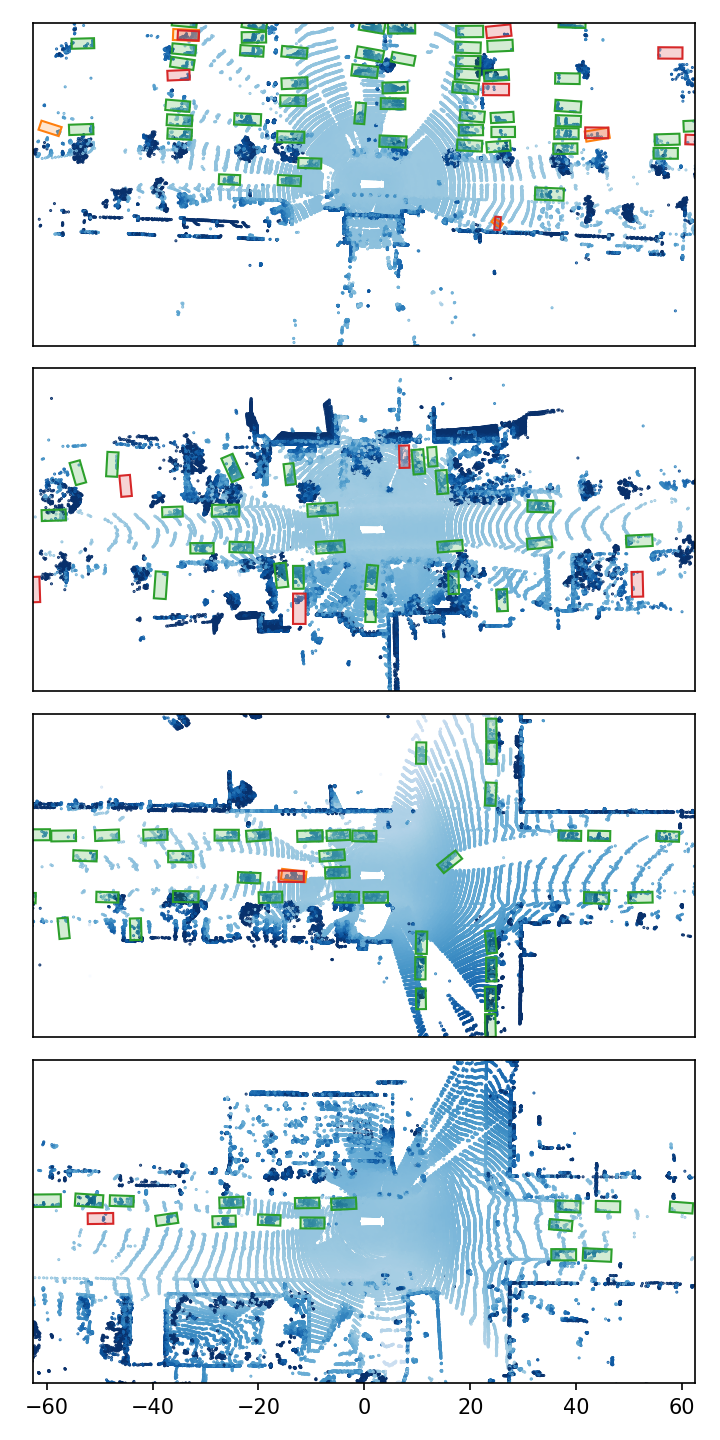}
        \caption{SOAP}
    \end{subfigure}
    \caption{Examples of pseudo-labels generated by different methods in nuScenes $\rightarrow$ Waymo semi-supervised domain adaptation setting. Green represents true positive pseudo-labels, orange represents false positive pseudo-labels, and red represents false negative pseudo-labels.}
    \label{fig:qualitative-ssda-nus-waymo}
\end{sidewaysfigure*}

We present qualitative results of SOAP pseudo-labels for nuScenes $\rightarrow$ Waymo in \cref{fig:qualitative-uda-nus-waymo,fig:qualitative-ssda-nus-waymo}. In both unsupervised and semi-supervised settings, we observe that SOAP pseudo-labels are more accurate compared to Direct, Co-training, and CutMix~\cite{Wang2022SSDA3DSD}, especially for far objects.

\vspace{2em}
\end{appendix}

 \clearpage
{\small
\bibliographystyle{ieee_fullname}
\bibliography{bib}
}

\end{document}